\documentclass[sigplan, nonacm]{acmart}
\usepackage[most]{tcolorbox}
\usepackage{booktabs}
\usepackage{subcaption} 
\usepackage{tikz}
\usepackage{soul}    
\usepackage{xcolor}
\usepackage[ruled,vlined]{algorithm2e}
\usepackage{ulem}

\newcommand{\filledcircled}[1]{%
\tikz[baseline=(char.base)]{
  \node[shape=circle, fill=black, text=white, inner sep=2pt] (char) {\small\bfseries #1};
}}
\AtBeginDocument{%
  }

\setcopyright{acmlicensed}
\copyrightyear{2018}
\acmYear{2018}
\acmDOI{XXXXXXX.XXXXXXX}
\acmConference[Conference acronym 'XX]{Make sure to enter the correct
  conference title from your rights confirmation email}{June 03--05,
  2018}{Woodstock, NY}
\acmISBN{978-1-4503-XXXX-X/2018/06}

\newcommand{\sys}{\textsc{NPUMoE}}
\begin{document}


\title{Efficient Mixture-of-Experts LLM Inference with Apple Silicon NPUs}

\author{Afsara Benazir}

\affiliation{%
  \institution{University of Virginia}
  \city{Charlottesville}
  \state{Virginia}
  \country{USA}
}
\email{hys4qm@virginia.edu}

\author{Felix Xiaozhu Lin}

\affiliation{%
  \institution{University of Virginia}
  \city{Charlottesville}
  \state{Virginia}
  \country{USA}
}
\email{felixlin@virginia.edu}

\looseness=-1
\begin{abstract} 
Apple Neural Engine (ANE) is a dedicated neural processing unit (NPU) present in every Apple Silicon
chip.
Mixture-of-Experts (MoE) LLMs improve inference efficiency via sparse activation but are challenging for NPUs in three ways: 
expert routing is unpredictable and introduces dynamic tensor shapes that conflict with the shape specific constraints of NPUs, 
several irregular operators e.g. top-k, scatter/gather etc. are not NPU-friendly, and launching many small expert kernels incur substantial dispatch and synchronization overhead. 
NPUs are designed to offload AI compute from CPU and GPU; our goal is to enable such offloading for MoE inference, particularly during prefill, where long-context workloads consume substantial system resources. 

This paper presents \sys{}, a runtime inference engine that accelerates MoE execution on Apple Silicon by offloading dense, static computation to NPU, while preserving a CPU/GPU fallback path for dynamic operations. 
\sys{} uses offline calibration to estimate expert capacity and popularity that drives three key techniques: (1) Static tiers for expert capacity to address dynamic expert routing (2) grouped expert execution 
to mitigate NPU concurrency limits (3) load aware expert compute graph residency 
to reduce CPU–NPU synchronization overhead. 
Experiments on Apple M-series devices using three representative MoE LLMs, four long-context workloads show that 
\sys{} consistently outperforms baselines reducing latency by 1.32x–5.55x, improving energy efficiency by 1.81x–7.37x, and reducing CPU-cycle usage by 1.78x–5.54x through effective NPU offloading. 










\end{abstract}
\maketitle

\section{Introduction}
\begin{figure}
    \centering
    \includegraphics[width=\linewidth]{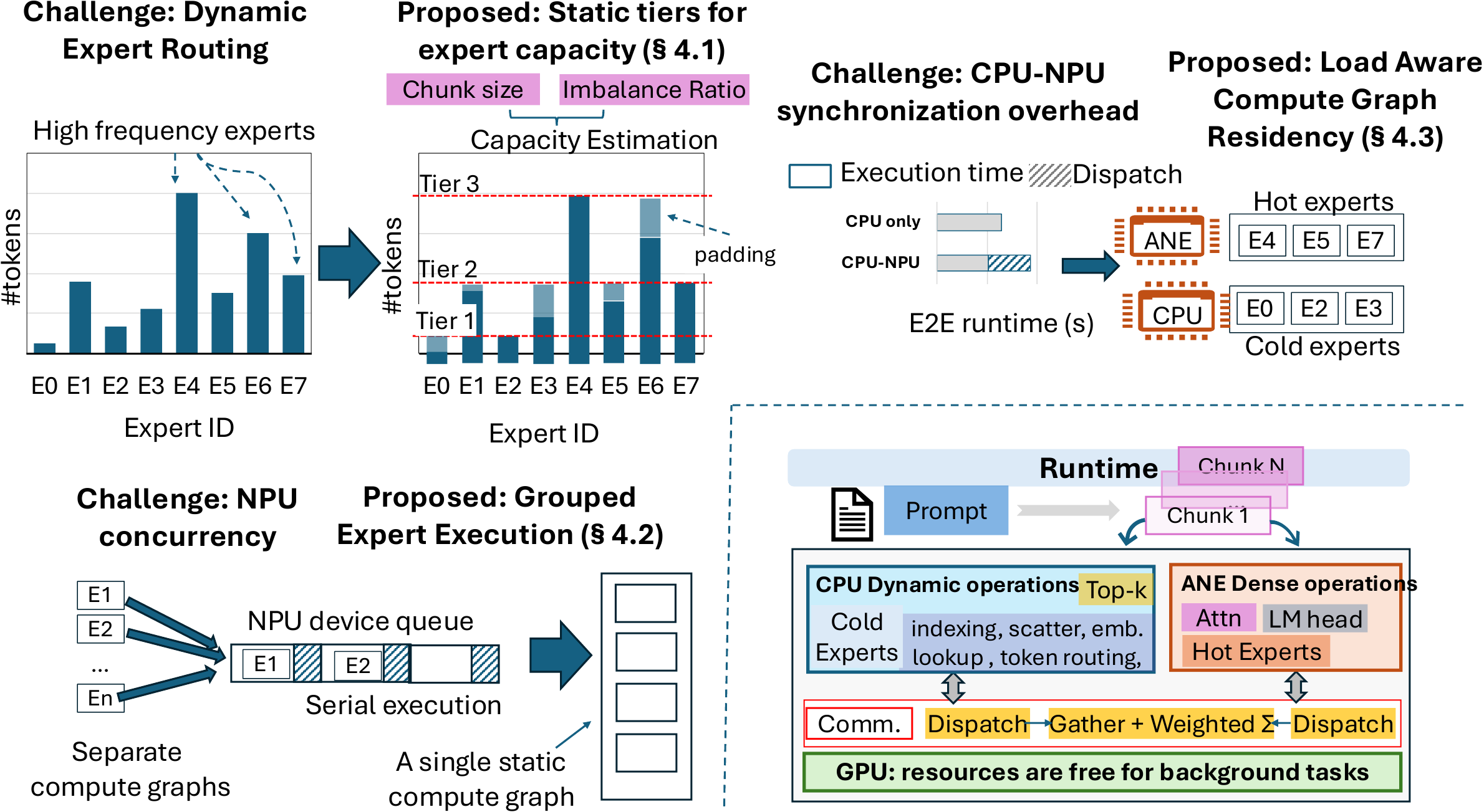}
    \caption{Our System: \sys{}}
    \Description{System overview showing the NPUMoE runtime pipeline, with token routing and runtime control on CPU, grouped expert execution on the Apple NPU, and GPU resources left available for concurrent background tasks.}
    \label{fig:o}
    \vspace{-5mm}
\end{figure}
\label{sec:i}

Recent advances in large language models (LLMs) increasingly adopt Mixture-of-Experts (MoE) architectures to scale model capacity efficiently through sparse activation \cite{shazeer2017outrageously}. On-device LLM inference ensures complete privacy and personalization but rarely runs in isolation: the CPU must keep the OS responsive (UI thread, I/O, networking etc.), and the GPU is often tasked with rendering/camera/display pipelines and interactive graphics.
Under heavy load, this can lead to resource contention and degrade system responsiveness \cite{wei2025agent}.

Neural processing units (NPUs) e.g Apple Neural Engine (ANE), Qualcomm Hexagon NPU, AMD NPUs etc. are widely integrated into commodity mobile platforms \cite{hao2025scaling}
offering high peak computing power
and increased power efficiency relative to CPUs/GPUs.
Prior work primarily focuses on efficient execution of dense transformer models on mobile GPUs \cite{cao2025moe} and NPUs \cite{xu2025fast, chen2025accelerating, chen2025heterollm}.
In contrast, MoE LLMs such as PhiMoE \cite{abdin2024phi}, Qwen3 MoE \cite{yang2025qwen3} etc. exhibit dynamic, sparse execution patterns that are fundamentally misaligned with NPU execution pipelines.


In addition to executing a dense attention block, in MoE architectures, a router dynamically chooses a subset of experts for each token instead of activating the entire network. Each expert is typically a small feed-forward block (FFN); tokens are routed to the selected experts for processing and then gathered back after computation. This introduces runtime dynamism characterized by dynamic expert selection, token-level routing, irregular memory access patterns, and varying load across different experts.


Apple Neural Engine (ANE) is Apple's dedicated hardware component for neural processing tasks to accelerate machine learning workload \cite{ane}, widely deployed in over two billion devices \cite{apple2023q1}. 
Under Apple’s unified memory architecture, the CPU, GPU, and ANE share a single memory pool; 
this allows large-capacity MoE models to remain fully resident on-device, reducing data movement overhead.
ANE is explicitly specialized for FP16 compute, with a 16-core ANE (Apple M2) delivering up to 15.8 TFLOPS compute \cite{ane_number}. These characteristics make Apple Silicon particularly attractive for on-device LLM inference at high precision.

\noindent\textbf{LLM acceleration on NPU.} 
NPUs support static, dense operations of typical LLMs, but its hardware design restricts execution of dynamic operations unique to MoE LLMs.
Offloading LLM compute to NPU offers key benefits: reduced resource contention - CPU/GPU can prioritize interactive tasks,
and keep other apps responsive; better power and energy efficiency \cite{ane_vision_tasks} 
which helps sustain performance longer on battery-powered devices. 
We identify a key \emph{opportunity} to adapt the dynamic shapes of MoE inference to the static execution constraints of NPU.

\begin{figure}[t]
    \centering
\includegraphics[width=0.8\linewidth]{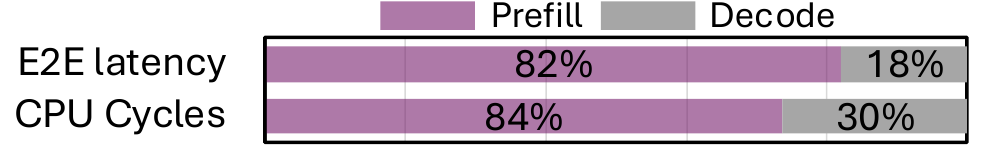}
    \caption{Prefill phase dominates end-to-end inference.}
    \Description{Bar chart comparing prefill and decode costs, showing that prefill accounts for most end-to-end latency and CPU-cycle consumption in long-context inference workloads.}
    \label{fig:prefill_dominate}
    \vspace{-5mm}
\end{figure}

During prefill, the context length is dynamic and depends on the specific prompt, whereas during decode only one token is processed owing to autoregressive generation. 
For long-context tasks, the prefill phase dominates inference \cite{jiang2024minference, du2025prefillonly}, accounting for 80\% of end-to-end inference latency and 70\% of CPU cycles (\autoref{fig:prefill_dominate}), thus we mainly target prefill acceleration in this work.


This work presents \sys{} (ref. \autoref{fig:o}), a runtime inference engine that accelerates MoE LLM inference on NPUs. 
We use Apple Neural Engine (ANE) as a representative mobile NPU to illustrate key characteristics and constraints of mobile NPUs due to its tight integration with Apple’s hardware-software stack, and widespread deployment across Apple devices. MoE execution on NPU presents several challenges.
\noindent\textit{\textbf{(1) Model dynamism complicates compute graph creation.}}
NPU is designed for static, shape-specific compute graphs (\S\ref{sec:b}) compiled ahead of time (e.g., via Core ML \cite{coreml}), that allows it to be compute efficient but conflicts with standard LLM inference where input tensors and operator shapes vary across prompts and batches.
Prior systems mitigate this via chunked prefill and graph bucketing \cite{agrawal2023sarathi, yin2025dynamic} but
such strategies presume dense, predictable structure and cannot be directly applied to all MoE operations. At runtime, each expert can get an arbitrary number of tokens and constructing graphs for all possible expert capacities is infeasible. 

\noindent\textit{\textbf{(2) Limited support for irregular/dynamic operators.}} 
NPU imposes strict constraints: (1) it is shape sensitive:  variable tensor shapes such as token to expert assignment, large dimension tensors in attention, unsupported operators etc. can trigger CPU fallback. (2) it is optimized for static dense operators and convolutions e.g. conv2D but cannot perform arbitrary runtime dependent core MoE operations such as dynamic indexing, tensor reshaping, top-k (router) etc. (3) Limited support for masking, reductions, and conditional logic complicates attention masking and expert aggregation.

\noindent\textit{\textbf{(3) Limited concurrency and frequent dispatch overhead.}}
NPUs have limited concurrency: many small compute graphs can serialize on a single device queue, causing high launch overhead and low utilization. 
Allocating individual compute graphs to each expert can cause a subgraph explosion, triggering CPU fallback.
CPU-NPU synchronization can account for >60\% of runtime in worst case (ref. \autoref{fig:cas}); 
this dominates end-to-end performance when a workload is split into many small subgraphs. Similarly, repeatedly invoking NPU for small kernels e.g., each expert's FFN can cause per-invocation overhead to dominate execution time.

To address these challenges, our \emph{insight} is to transform the MoE execution pipeline to maximize NPU utilization 
while retaining CPU/GPU fallback for dynamic operations.
\noindent\textbf{Goal.}
Our overarching principle is straightforward: while it would be ideal to execute all computations on NPU, MoE dynamism makes this impractical; therefore, we aim to maximize NPU offload/usage while choosing which parameters and operations to offload and which compute unit to utilize through careful graph design. 



\noindent\textbf{Design (\S\ref{sec:overview})} 
Our design treats NPU as the primary execution engine; CPU/GPU acts as a secondary fallback path for unsupported operations.
Dense projections such as attention and expert FFN blocks are primarily executed on NPU, dynamic operators whose behavior depends on runtime input such as top-k, dynamic indexing etc. are delegated to CPU/GPU. Expert activations are returned to CPU for routing, aggregation, and fallback execution when needed.

We, thus (1) partition runtime operators across compute units that maximize energy efficiency and overall utilization while meeting latency targets, (2) construct efficient NPU-friendly compute graphs employing non-trivial techniques: 
\noindent\textbf{Static Tiers for Expert Capacity (\S\ref{sec:sect})}
\sys{} respects NPU shape sensitivity:
each expert is bounded to a fixed token capacity per layer. 
This introduces an explicit capacity/accuracy tradeoff: smaller capacities reduce wasted compute and improve utilization (less padding),
but increase the likelihood of overflow/token dropping when many tokens route to the same expert, affecting accuracy. 
\sys{} uses offline calibration to estimate each expert’s expected workload and maps experts to static capacity tiers accordingly: popular experts receive larger capacities (e.g., top quartile gets 4x base capacity), while less frequently used experts receive smaller capacities, focusing resources on experts that most impact compute and latency. 


\noindent\textbf{Grouped Expert Execution (\S\ref{sec:gee})}
\sys{} accommodates NPU concurrency limitations: fine-grained expert execution creates too many small compute graph launches that can serialize on the same device queue, increasing launch count and data movement overheads.
To mitigate serialization of compute graphs,
\sys{} batches multiple experts into a single static, grouped dense FFN compute graph, so that one invocation executes several experts as a single fused block with shared launch overhead and scheduling costs, thus maximizing NPU utilization.
Grouping creates an explicit tradeoff between group size and per-expert capacity.
\noindent\textbf{Load Aware Expert Compute Graph Residency (\S\ref{sec:lacgr})}
CPU/NPU synchronization introduces latency; \sys{} uses load-aware placement of grouped expert compute graphs across available execution units to meet latency targets. In particular, (1) short, bursty workloads - such as cold experts that receive only a few routed tokens underutilize NPU and  increase end-to-end latency and (2) small groups of few experts 
cannot amortize graph launch overhead. 
In such cases, running them on CPU avoids this dispatch overhead. \sys{} therefore maintains a small resident \textit{working set} of expert compute graphs on NPU, formed by hot expert groups whose launches can be amortized, while relegating cold experts to the CPU/GPU fallback path. 
Expert popularity (\S\ref{sec:lacgr})
determines the hot/cold state of each expert.


\noindent\textbf{Implementation and Evaluation (\S\ref{sec:eval})}
We implement \sys{} from scratch, using the ANEMLL \cite{anemll} library for model conversion and Apple’s Core ML \cite{coreml} stack (via Swift) for runtime. We compare against three baselines (in \S\ref{sec:eval}) that differ in how MoE runtime operations are executed, with primary emphasis on long-context prefill workloads.
We evaluate \sys{} on three popular MoE LLMs: Phi-3.5-MoE-Instruct, Phi-tiny-MoE-Instruct, and Qwen3-30B-A3B across two Apple M-series devices: M2 Ultra and M2 Max.

\sys{} consistently provides the best performance-efficiency tradeoff across workloads: it reduces prefill latency by 1.32x–5.55x, energy consumption by 1.81x–7.37x and CPU-cycle usage by 1.78x–5.54x, with negligible accuracy loss (<1.1\% compared to FP16) on our test datasets. These gains come from our proposed techniques (\S\ref{sec:sect}-\S\ref{sec:lacgr}) that accelerate the model execution path through effective offloading of compute to NPU and from carefully balancing grouping, capacity selection, and padding overhead. Overall results demonstrate the effectiveness of exploiting mobile NPUs for dynamic MoE model inference. We will make our codebase open-source.
Our contributions are as follows:
\begin{itemize}
 \item We investigate the challenges and limitations of MoE inference on NPUs and perform an in-depth analysis of MoE on Apple Neural Engine.
   
    \item We introduce three novel techniques: static tiers for expert capacity, grouped expert execution, and load-aware expert residency. 
    Combined, they accelerate MoE inference and efficiency significantly.
    
    
    

    \item We design an inference engine \sys{}, 
    a first-of-its-kind system that systematically optimizes MoE efficiency for NPUs. 
    Our comprehensive end-to-end evaluation demonstrate the superiority of \sys{} over competitive baselines. 
    
\end{itemize}





\section{Background}
\label{sec:b}

\begin{figure}[t]
    \centering
    \includegraphics[width=0.5\linewidth]{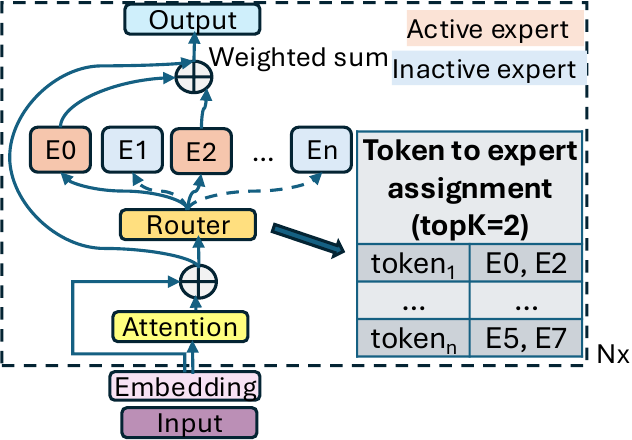}
    \caption{Mixture of Experts architecture}
    \Description{Diagram of a mixture-of-experts layer in which a router selects a small subset of experts for each token and merges their outputs back into the main model flow.}
    \label{fig:moe}
\end{figure}

\noindent\textbf{Mixture of experts.}
architecture popularized by models such as Switch Transformer \cite{fedus2022switch}, GShard \cite{lepikhin2020gshard} etc. performs sparse computation by routing tokens to specialized feed forward network (FFN) blocks (i.e. experts) (ref. \autoref{fig:moe}).
MoE inference is fundamentally dynamic: the router performs top-k selection and dispatches tokens via scatter, performs dense expert FFN compute, gathers expert output, and attention and each expert must handle variable input tokens.
\begin{figure}[t]
    \centering
    \includegraphics[width=0.8\linewidth]{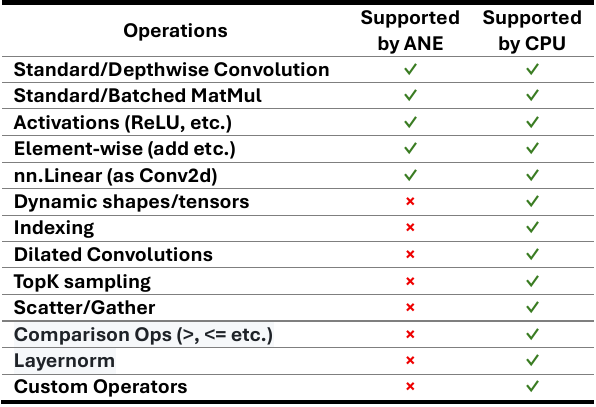}
    \caption{Runtime operations supported by ANE and CPU.}
    \Description{Comparison of MoE runtime operations, indicating which ones can run efficiently on the Apple Neural Engine and which ones remain on the CPU.}
    \label{fig:aso}
\end{figure}

\noindent\textbf{Apple Neural Engine.} is Apple Silicon’s dedicated NPU for energy-efficient, high-throughput neural network inference. Its unified-memory architecture allows the CPU, GPU, and ANE to access the same memory pool, making the platform attractive for on-device LLM inference; e.g., M2 Ultra with 192GB RAM integrates a 32-core Neural Engine rated at up to 31.6 TOPS \cite{ane}. 
However, similar to other commodity NPUs, ANE imposes static graph and shape rigidity constraints and offers limited support for runtime dynamic operations \cite{xu2025fast, yin2025dynamic}; list of supported operations in \autoref{fig:aso}. 
\textit{Terminology}: Core ML \cite{coreml} is Apple’s machine learning framework optimized 
to minimize memory/power consumption.

\begin{figure}[t]
    \centering
    \includegraphics[width=\linewidth]
    {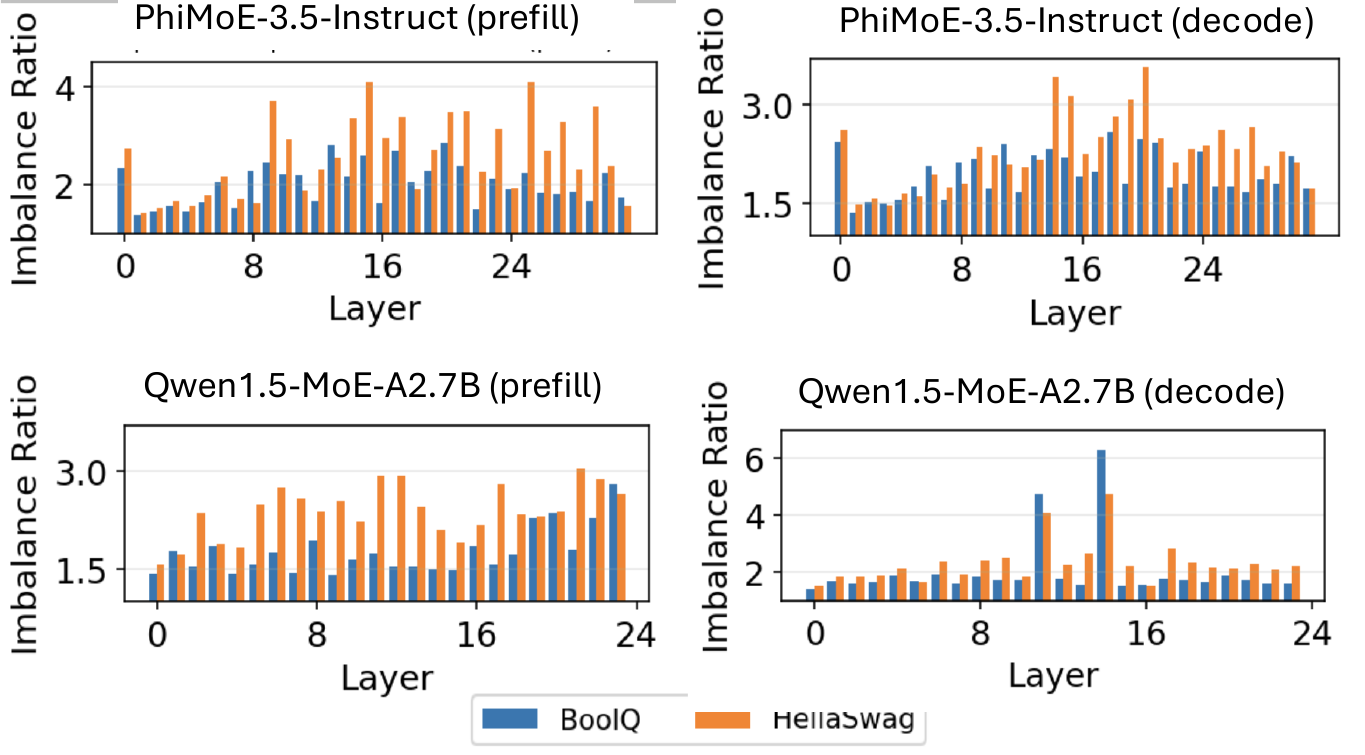}
    \caption{Expert load imbalance in prefill and decode for Phi-3.5-MoE and Qwen1.5-MoE-A2.7B. Imbalance ratio is the maximum expert load divided by the average layer load; values above 1 indicate greater skew.}
    \Description{Plots of expert-load imbalance for two MoE models during prefill and decode, showing that routed-token demand is highly skewed across experts and varies by execution phase.}
    \label{fig:eli}
\end{figure}

\noindent\textbf{Motivation.} 
Expert routing is highly imbalanced \cite{fedus2022switch}: a small subset of experts receive most tokens, while others are rarely activated, leading to wasted compute, extra communication and memory traffic, and throughput bottlenecks (\S\ref{sec:b}). 
Prior work \cite{fedus2022switch, liu2024deepseek, abdin2024phi, liu2024deepseek_v3} addresses this issue mainly during training through auxiliary load-balancing losses and heuristics to ensure  uniform distribution of tokens across all experts.
In contrast, our microbenchmark (\autoref{fig:eli}) show that expert load during inference remains highly skewed and varies across layers. 
This motivates treating expert execution as a load-aware systems problem, where compute placement and expert capacity are tailored to expected workload.

\begin{figure}
    \centering
    \includegraphics[width=0.9\linewidth]{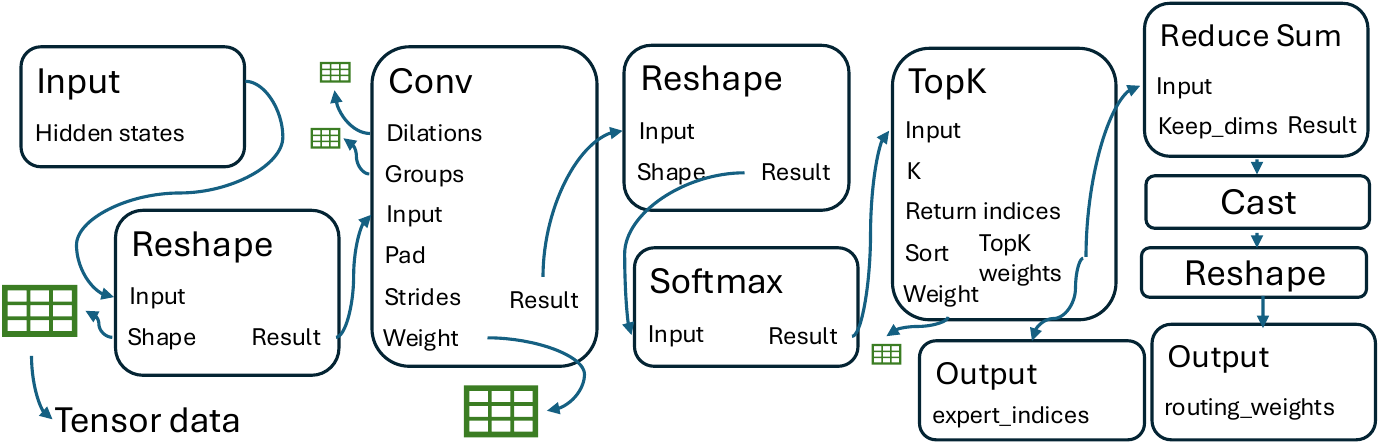}
    \caption{A compute graph of a single MoE router.}
    \Description{Example compute graph for a single MoE router, illustrating the sequence of routing, packing, expert execution, and merge operations.}
    \label{fig:cg}
\end{figure}
\noindent\textbf{Compute graph.} is a sequence of connected operations that define data flow through the model (ref. \autoref{fig:cg}). Prior work \cite{chen2025heterollm} and \S\ref{app:cg} show that graph build time is significantly higher than execution time, making runtime dynamic graph construction on NPU impractical.
Each compute graph is unique, components with similar shape but different weights cannot trivially share a single graph \cite{coreml}.

\noindent\textbf{Scenario.} Two deployment scenarios exist: (1) Available memory is limited, weights are swapped in and out from host/flash memory at runtime (e.g. non-unified memory devices) and (2) Available memory large enough to hold all experts (e.g. unified memory of Apple Silicon). Prior work \cite{cao2025moe} has focused mostly on scenario 1 and adopted techniques such as expert prefetching and caching. 
We focus on scenario 2 and assume all model weights reside in device memory.

\begin{figure}
    \centering
    \includegraphics[width=0.8\linewidth]{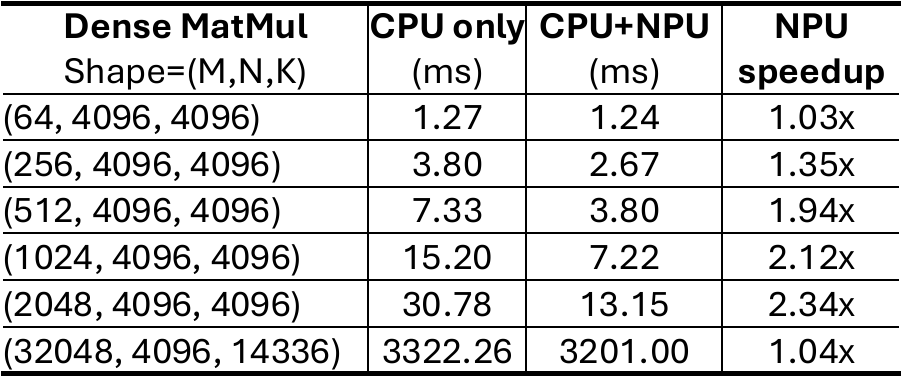}
    \caption{Runtime of dense matmuls on M2 Ultra demonstrates NPU efficiency. Benefit is reduced for very small (M=64) or very large matrices (M=32048).}
    \Description{Performance plot for dense matrix multiplications on M2 Ultra showing that the NPU is most efficient at mid-sized workloads, with smaller gains for very small or very large matrices.}
    \label{fig:ne}
\end{figure}
\noindent\textbf{NPU Efficiency.}
We microbenchmark several dense matmul operations on M2 Ultra to compare NPU and CPU efficiency (ref. \autoref{fig:ne}); across varying input shapes NPU demonstrates 1.4x–2.3x lower runtime than CPU. However, this benefit appears only when enough work is amortized per launch and graph compute isn't too small (M=64) or too large (M=32048 or vocab size).

\begin{figure}
    \centering
    \includegraphics[width=0.7\linewidth]{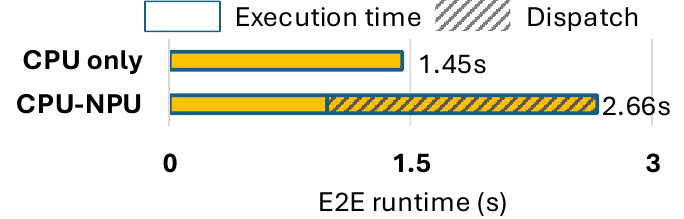}
    \caption{CPU-NPU synchronization overhead.}
    \Description{Plot showing how additional CPU-to-NPU synchronization overhead increases end-to-end runtime, illustrating the cost of frequent accelerator coordination.}
    \label{fig:cas}
    \vspace{-5mm}
\end{figure}
\noindent\textbf{CPU-NPU coordination.} pays the cost of high synchronization \cite{dai2026accelerating}. Unified memory helps communication, but it does not eliminate coordination cost; several recent systems show that synchronization and launch overheads can be comparable to kernel execution time for fine-grained workloads \cite{chen2025heterollm}. 
As shown in \autoref{fig:cas}, frequent short NPU invocations amplify dispatch overhead and device-queue serialization, reducing overall NPU utilization.





\noindent\textbf{Why not GPU by default?} For on-device mobile LLM inference, the GPU should not be assumed to be the default accelerator as it is a shared system resource tasked with graphics rendering, display, and other latency-sensitive interactive workloads \cite{yi2020heimdall} whereas ANE is a dedicated ML accelerator designed for sustained throughput and energy efficiency \cite{hubner2025apple}.
For long-context workload, prefill dominates end-to-end inference \cite{agrawal2023sarathi} thus accelerating dense prefill for NPU is more valuable than monopolizing GPU for compute. Apple advocates harnessing ANE for demanding ML workload leaving GPU to execute non-ML workloads \cite{ane_number, apple2024visiontransformersane}. 

\begin{figure*}
    \centering
    \includegraphics[width=0.9\linewidth]{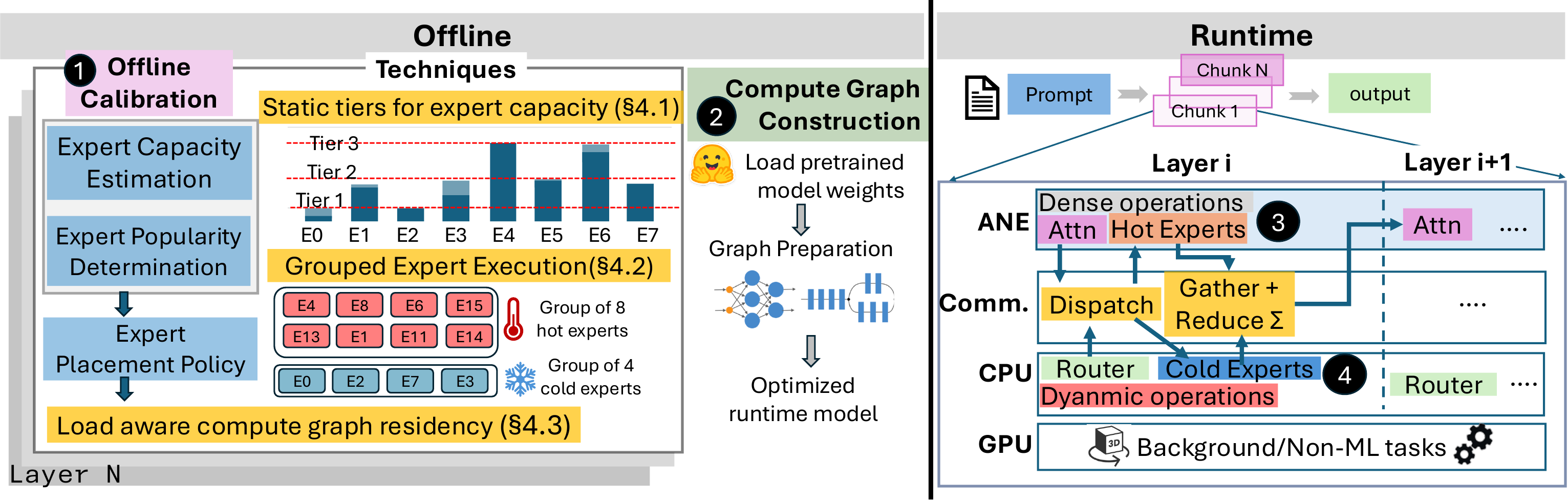}
    \caption{Our system \sys{} - incorporates key design (1) Static tiers for expert capacity (\S\ref{sec:sect}) (2) Grouped Expert Execution (\S\ref{sec:gee}) and (3) Load aware compute graph residency (\S\ref{sec:lacgr}). GPU resources remain available for background tasks while CPU and NPU synchronize between runtime operations.}
    \Description{End-to-end system diagram for NPUMoE showing static capacity tiers, grouped expert execution, and load-aware graph residency across CPU, NPU, and GPU resources.}
    \label{fig:system}
\end{figure*}
\section{\sys{}: Overview}
\label{sec:overview}



\subsection{Workflow}
\noindent\textbf{Workflow} consists of two phases (ref. \autoref{fig:system}):

\noindent\textbf{Offline}: \filledcircled{1}
We calibrate experts offline to estimate expert capacity tiers (\S\ref{sec:sect}) and expert popularity to derive an optimal expert placement policy (\S\ref{sec:lacgr}).
\filledcircled{2} Using this calibration data, we apply our key techniques (\S\ref{sec:sect}-\S\ref{sec:lacgr}) to construct the optimized Core ML compute graph: pretrained model weights are loaded, the optimized graph is prepared for conversion, and compiled for its target compute units.

\noindent\textbf{Runtime}: \filledcircled{3} When a prompt arrives, we process the prefill chunk and offload NPU-supported static, dense operations (i.e. attention and expert FFN) for that chunk; activations return to CPU. \filledcircled{4} Remaining dynamic operations follow the fallback path: routing, top-k dispatch etc. runs on CPU, 
selected (hot) experts execute on NPU while cold experts fallback to CPU. 
CPU and NPU synchronize during execution, while GPU resources remain available for background tasks.

\subsection{Design Principle}
\label{sec:dp}
At a high level our target principle is (1) to maximize compute offloaded to NPU while treating the CPU (or GPU) as a fallback path for dynamic or unsupported operations.
and (2) optimize for resource efficiency. In particular, we seek to achieve target latency goals, improve energy efficiency, reduce CPU cycles used and minimize CPU–NPU synchronization.
Variable prompt length breaks static attention graphs; padding to max context (e.g. 128K) is infeasible, hence following prior work we adopt chunked prefill \cite{xu2025fast}.



\noindent\textbf{Operator Partitioning}
We employ a hybrid partitioning strategy that ensures maximum utilization of available resources respecting target latency/energy.
We place dense workloads on NPU, while keeping dynamic or control-heavy operations on CPU. Specifically, attention (Q/K/V projections, softmax, stateful KV cache update) and expert FFN compute execute on NPU with static tensor shapes, whereas layerNorm, masking, expert routing (top-$k$/argmax), and token dispatch/merge operations (scatter, gather, weighted combine) execute on the CPU due to their dynamic indexing and irregular memory access patterns.

\noindent\textbf{Considerations for graph construction}
Aside from static shapes, efficient NPU execution also depends on tensor layout. Shapes that are powers of two and sizes that are multiples of 16 (e.g., 16, 32, 48, 64) improve memory access and utilization \cite{antmikinka2024ane}; tensors should be 16-byte aligned to ensure efficient execution and compatibility with NPU hardware constraints. In practice, transformer linear layers (\texttt{nn.Linear}) are often rewritten as 1x1 convolutions (\texttt{nn.Conv2d}) to better utilize NPU-optimized kernels \cite{ane_number},  fixed-shape KV caches are also standard implementation choices for Core ML \cite{anemll}.

\noindent\textbf{Optimization}
When routed tokens exceed an expert’s static capacity (\S\ref{sec:sect}), \sys{} prunes the overflow tokens using an activation-based saliency score to limit accuracy degradation: tokens with lower saliency, ranked by the L2 norm of attention output activations are discarded first, following prior activation-magnitude pruning methods ~\cite{liu2023deja, akhauri2024shadowllm}. The saliency score is derived from activations, while overflow ranking and pruning are performed on CPU. Inference accuracy is reported in \autoref{sec:eval}.


\subsection{Offline Calibration}
\label{sec:oc}
We perform an offline calibration pass that measures per-layer expert routing statistics and derives an expert popularity ranking. \sys{} then uses this ranking to configure two runtime decisions: capacity-tier assignment (\S4.1) and expert compute graph residency (\S4.3).
For each layer, we record how often each expert is selected and rank experts by their cumulative token assignments. Details in \autoref{sec:eval}.
We validate our profiling along two dimensions: (1) accuracy of expert popularity estimation, and (2) cross-dataset stability of dominant experts in \S\ref{sec:as}.





\section{System Design}
\noindent\subsection{Static Tiers for Expert Capacity}
\label{sec:sect}
\begin{figure}[t]
    \centering
    \includegraphics[width=\linewidth]{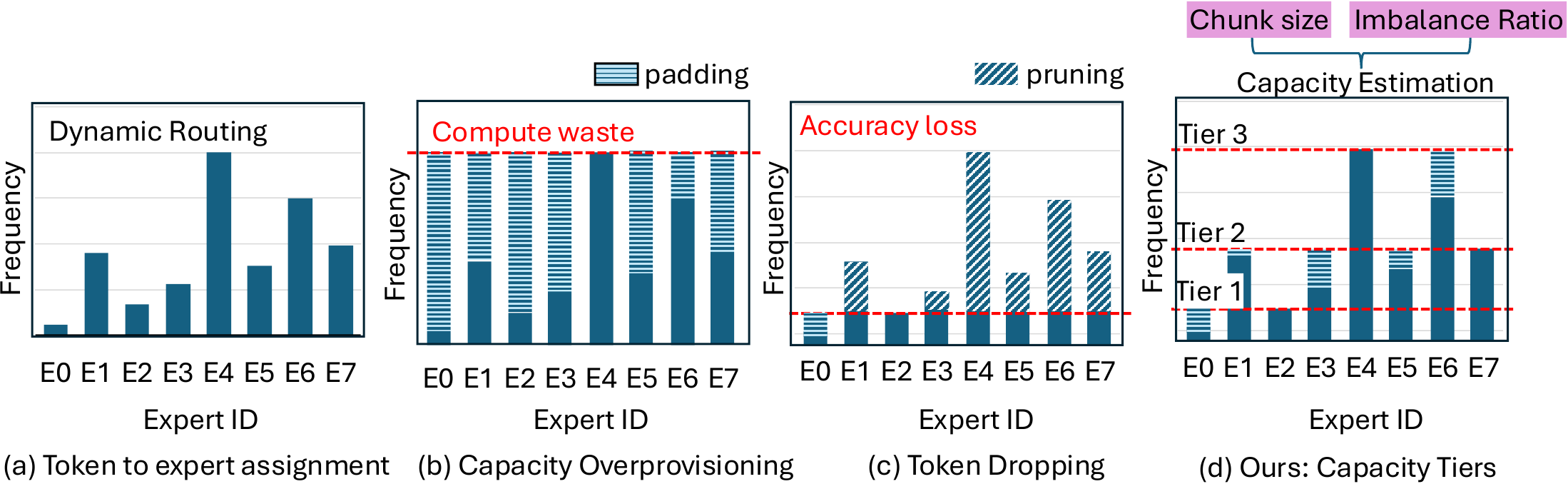}
    \caption{Static tiers for expert capacity. (\S\ref{sec:sect})}
    \Description{Illustration of static expert-capacity tiers used to bucket routed-token counts into a small set of fixed-capacity execution graphs.}
    \label{fig:ect}
\end{figure}
\textbf{Issue}: A key challenge in MoE inference is that token-to-expert assignment is inherently unpredictable: after routing, different experts receive variable numbers of tokens, and this load can vary across prompts and layers. This dictates variable length input to each expert, hence dynamic input shapes and dynamic compute graphs at runtime; such variability interacts poorly with NPU execution, which cannot execute dynamic shapes at runtime.


\noindent\textbf{Approach}: To make expert execution compatible with NPU, we replace dynamic per-expert token counts with estimated fixed capacities proportional to each expert’s expected load.
When an expert receives more tokens than its capacity, two overflow policies are possible:
(i) \emph{padding}, i.e., allocating a larger uniform capacity to all experts to avoid dropping tokens, at the cost of lower utilization; or
(ii) \emph{pruning}, which drops a small fraction of routed tokens based on our optimization in \S\ref{sec:dp}.



\noindent\textbf{Problem Formulation}
\autoref{fig:ect} illustrates a tiered capacity design that converts highly skewed token-to-expert assignment into a small number of static shapes.

Let $n_e$ be the number of tokens routed to expert $e$ in a layer after top-$k$ routing. Because $\{n_e\}_{e=1}^{E}$ is typically imbalanced (ref. \autoref{fig:ect}a), provisioning all experts for the worst case,
$
C_e = \max_{e'} n_{e'},
$
avoids overflow but incurs substantial padding cost (ref. \autoref{fig:ect}b). Using a smaller uniform capacity $C$ reduces padding, but when $n_e > C$, the overflow
$
d_e = \max(0, n_e - C)
$
must be dropped (fig.\ref{fig:ect}c), degrading overall accuracy. A uniform fixed-capacity design therefore trades off compute waste against accuracy loss.

To avoid this, we organize capacities into a small number of tiers,
$
T = \{C^{(1)}, C^{(2)}, C^{(3)}\}, \quad C^{(1)} > C^{(2)} > C^{(3)},
$
and assign each expert to the smallest tier that matches its expected load:
$
\tau(e) = \max \{j \mid n_e \le C^{(j)}\}.
$
High frequency experts are mapped to higher-capacity tiers, while low frequency experts use smaller tiers (fig.\ref{fig:ect}d).



\noindent\textbf{Expert Capacity Estimation.}
We use the prefill chunk size $B$ and the per-layer routing imbalance ratio $r_\ell$ to estimate expert capacity at layer $\ell$. As shown in \autoref{fig:eli}, we first measure the expected token load of each expert from a calibration dataset and compute the imbalance ratio as
\[
r_\ell = \frac{\max_{e \in \{1,\dots,E\}} n_{\ell,e}}{\frac{1}{E}\sum_{e=1}^{E} n_{\ell,e}},
\]
where $n_{\ell,e}$ denotes the number of tokens routed to expert $e$ at layer $\ell$, and $E$ is the number of experts in that layer. Thus, $r_\ell$ is computed per layer and captures routing skew: $r_\ell=1$ indicates perfectly balanced routing, while larger values indicate greater imbalance.

We define the base capacity as \(C_{\text{base}}=B/E\), i.e., the average per-expert load under balanced routing. 
Using $r_\ell$, the busiest expert at layer $\ell$ is estimated as
\[
n_{\ell,\max} \approx r_\ell \cdot \frac{B}{E}.
\]
For example, if $B=256$, $E=8$, and $r_\ell=2$, then
$
n_{\ell,\max} \approx 2 \cdot \frac{256}{8} = 64
$
tokens. Layers with higher routing skew use larger capacities, while more balanced layers use smaller ones to reduce padding.

For completeness, overflow tokens cannot be easily spilled to CPU/GPU as Core ML determines the compute graph and device placement statically at compile time. Dynamic tensor partitioning across devices is therefore unsupported, making static capacity estimation the more practical design choice. 





\subsection{Grouped Expert Execution}
\label{sec:gee}
\begin{figure}
    \centering
    \includegraphics[width=\linewidth]{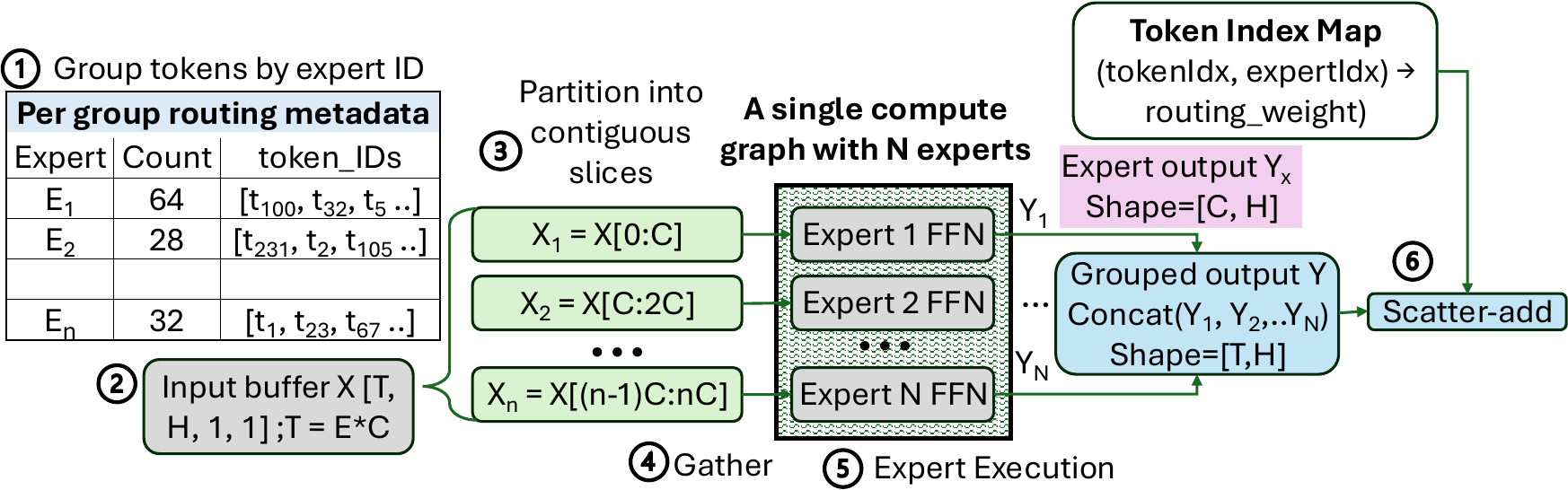}
    \caption{Grouped Expert Execution (\S\ref{sec:gee})}
    \Description{Illustration of grouped expert execution in which routed tokens are packed into fixed-capacity expert slices, executed together, and then scattered back to their original order.}
    \label{fig:gee}
\end{figure}
Following \S\ref{sec:sect}, a naive, straightforward approach is to allocate a separate compute graph for each expert after estimating its capacity but this incurs non-trivial dispatch overhead. 

\noindent\textbf{Issue}: Fine-grained per-expert execution launches many small graphs at runtime. Since these launches typically share a single device queue, they are serviced sequentially, limiting NPU concurrency.

Concurrent host threads may submit work in parallel on NPU, but that does not guarantee concurrent execution on the accelerator \cite{xue2023v10, xue2024hardware}.
When each graph is short-lived (such as a single expert), 
fixed launch overhead can dominate latency; launching all  experts separately may cost more than the useful NPU compute itself \cite{aimuyo2025flashmoe}.


\noindent\textbf{Approach:}
\sys{} groups multiple experts into a dense, static FFN compute graph, enabling a single invocation to process several experts as a fused block with shared launch and scheduling overhead. By consolidating experts within a unified compute graph, \sys{} reduces the overhead associated with individual launches. Each expert in a single group have the same capacity tier to prevent dynamic shapes.





\noindent Execution illustrated in \autoref{fig:gee} is as follows.
\noindent\filledcircled{1} \textbf{Group tokens by expert ID.} For a group of \(G\) experts with per-expert capacity \(C\), the runtime first determines which routed tokens belong to each expert $e_g$ in that group.
\noindent\filledcircled{2} \textbf{Allocate a grouped input buffer.}
$
X \in \mathbb{R}^{(G \cdot C)\times H \times 1 \times 1},
$
where \(H\) is the hidden dimension. 
\noindent\filledcircled{3} \textbf{Partition into contiguous expert slices.} The buffer
is viewed as \(G\) contiguous expert slices
\[
X = [X_1; X_2; \dots; X_G], \qquad X_g \in \mathbb{R}^{C \times H \times 1 \times 1}.
\]

Here, \(X_g\) contains the packed token block for expert \(e_g\). 
\noindent\filledcircled{4} \textbf{Gather routed tokens into expert slices.} Let \(n_g\) denote the number of routed tokens assigned to expert \(e_g\), with \(n_g \le C\). The routed token embeddings \(T_g \in \mathbb{R}^{n_g \times H \times 1 \times 1}\) are packed into the front of slice \(X_g\), while the remaining rows are zero-padded:
$
X_g[0:n_g] \leftarrow T_g, \quad X_g[n_g:C] \leftarrow 0.
$

\noindent\filledcircled{5} \textbf{Expert execution.} Each expert operates
on its own static capacity slice,
$Y_g = f_{e_g}(X_g),$
and the grouped output is formed by concatenation:
$
Y = \mathrm{concat}(Y_1,\dots,Y_G).
$

\noindent\filledcircled{6} \textbf{Weighted scatter.}
After execution, the runtime scatters only the valid first $n_g$ rows from each output slice, discards padded rows, and cumulates the results back to their original token position multiplied by their corresponding routing weights.




This layout eliminates dynamic indexing inside the compiled graph: expert-to-token assignment is resolved before invocation, and the model only sees a dense, statically shaped tensor.
For example, if four experts receive \((64,32,32,0)\) tokens and \(C=64\), then the first slice is fully occupied, the next two are partially padded, and the last slice is entirely padding. This converts imbalanced routing into a dense, static layout suitable for grouped execution.

\noindent\textbf{Configuration:} We explore configurations with different group sizes (e.g., 4, 8, and 16 experts per graph) and fixed per-expert capacities (e.g., 16, 32, and 64). These choices create explicit trade-offs:
(1) \textit{Large group size + higher capacity (e.g., $8\times64$)} improves launch amortization and reduces queue pressure. However, it can overprovision lightly used experts and waste compute; in extreme cases, it can block group parallelism and fall back to CPU execution (e.g., $16\times128$).
(2) \textit{Large group size + lower capacity ($8\times16$)} still benefits from launch amortization with less wasted compute, but overflow risk rises under bursty workloads.
(3) \textit{Small group size + higher capacity ($4\times64$)} handles imbalanced routing better than larger groups and reduces overflows, but it requires more launches and can still waste work through padding.
(4) \textit{Small group size + lower capacity ($4\times16$)} is efficient under light, balanced loads, but it needs frequent dispatches and can still face token overflow.

\noindent\textbf{Observation}: We observe a hard feasibility boundary:
if a compute graph becomes too large (e.g. >1.2 GB for M2 Ultra),
execution falls back to CPU;
grouping all 16 experts of PhiMoE into a single graph is not optimal for NPU; careful consideration is needed as
to how much weight to accumulate in one compute graph. Impact of grouping is in \S\ref{sec:single_moe_block}.



\subsection{Load-Aware Compute Graph Residency}
\label{sec:lacgr}
\begin{figure}
    \centering
    \includegraphics[width=\linewidth]{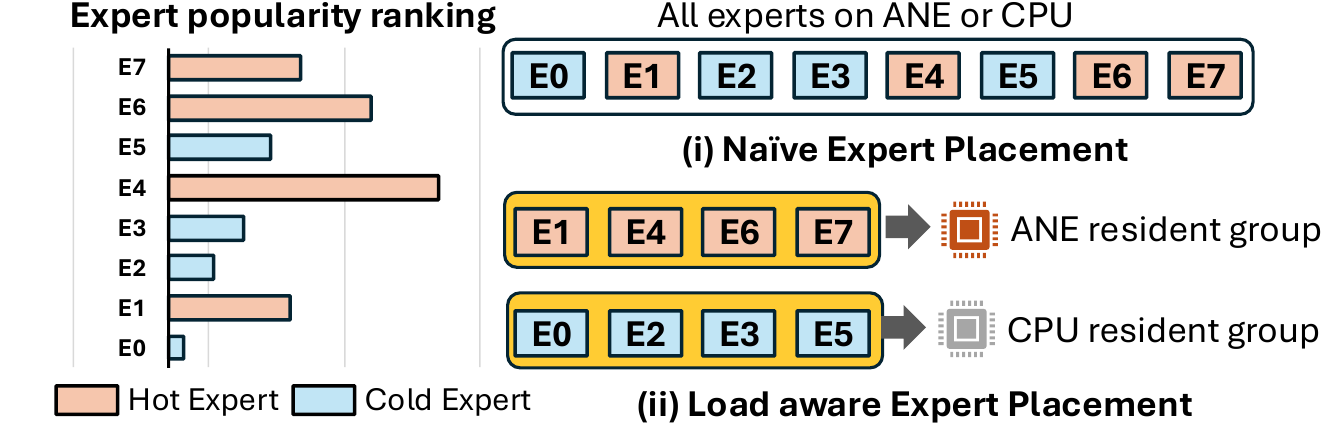}
    \caption{Expert placement based on expert popularity on different compute units (\S\ref{sec:lacgr}). E0 means expert 0 and so on.}
    \Description{Expert-placement visualization showing how hot experts are mapped to NPU-resident grouped graphs while colder experts remain on CPU or GPU fallback paths.}
    \label{fig:expert_placement}
\end{figure}
\textbf{Issue 1}: Grouped expert execution amortizes launch overhead, but the choice of grouping interacts directly with routed token load. When a high-demand expert is grouped with low-demand ones, the system must either overprovision the whole group through padding to the highest tier in that group, causing poor utilization, or keep capacity low and force the hot expert to drop tokens.

\noindent\textbf{Issue 2}:  
Uniform NPU offload is not always beneficial as CPU-NPU coordination introduces latency.
Small, bursty workloads caused by cold experts or very fine-grained grouping often fail to amortize CPU-NPU dispatch and synchronization overhead; this launch cost can outweigh useful compute and increase end-to-end latency. 



\noindent\textbf{Approach:} \sys{} uses a load-aware expert placement policy that jointly decides which experts are grouped together and where the resulting grouped compute graph should reside. The policy is driven by offline calibration, which estimates per-layer expert popularity from routing statistics, and by grouping granularity, which determines whether a grouped configuration contains enough useful work to amortize graph launch overhead.
\sys{} keeps a small resident working set of hot (high-demand) experts on NPU to amortize dispatch overhead, while 
configurations with fine-grained grouping (such as $2\times8$), where individual groups are small and bursty, are relegated to the CPU/GPU fallback path.
This avoids unnecessary CPU-NPU synchronization for small workloads, reduces compute waste, and improves utilization by matching expert demand to the most suitable execution unit. 

\noindent\textbf{Expert Popularity.}
Offline calibration ranks experts at each layer by cumulative routed-token count, revealing a skewed activation pattern in which a small subset of experts consistently receives most routed tokens. We treat these as hot experts and the remainder as cold experts. 
We find that dominant experts are not strongly dataset-specific (ref. \S\ref{sec:as}): experts that are highly active on one benchmark tend to remain highly active on another. 

\noindent\textbf{Grouping granularity.}
\sys{} groups experts by expected demand: hot experts are grouped together, while cold experts are grouped separately. This avoids mixing highly active and rarely used experts in the same graph, which would otherwise increase padding and reduce useful work per invocation. Impact of group size is in \S\ref{sec:single_moe_block}.

\noindent\textbf{Expert Compute Graph Placement Policy} illustrated in \autoref{fig:expert_placement}, determines which experts should be grouped and, ultimately, where their compute graphs should reside. 
\sys{} assigns grouped expert graphs to the execution unit that best matches their expected load. Hot groups remain on NPU because their demand is sufficient to amortize launch and synchronization cost, whereas cold groups or very small groups are mapped to CPU, where short, bursty workloads avoid NPU dispatch cost. This reduces unnecessary CPU-NPU coordination and improves NPU efficiency.

\section{Evaluation}
\label{sec:eval}
\subsection{Experimental Setup}
\label{sec:exp_setup}
\noindent\textbf{Hardware}
Our primary target platform is Apple M-series devices representing unified memory architecture. 
We use an M2 Max (64GB memory, 12-core CPU, 16-core ANE) and an M2 Ultra (192GB memory, 24-core CPU, 32-core ANE) as our hardware testbed. 




\begin{table}
    \centering
    
    \caption{Models used in experiment.}
    \vspace{-2mm}
    \includegraphics[width=\linewidth]{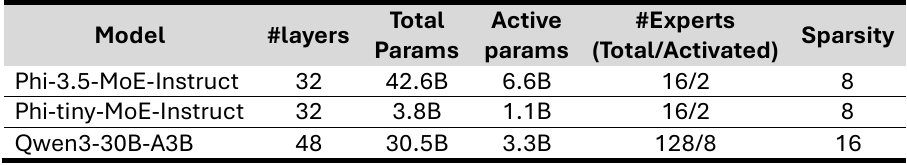}
    \Description{Table summarizing the evaluated MoE models, including model family, parameter scale, expert count, routing configuration, and model size.}
    \label{tab:model}
\end{table}
\noindent\textbf{Model}
We experiment with three popular MoE models as illustrated in \autoref{tab:model} representing state-of-the-art (SOTA) MoE models with expert counts ranging from 16 to 128 (1) Phi-3.5-MoE-instruct \cite{microsoft2024phi35moeinstruct}, a 32-layer MoE model with 16 experts and top-k=2 routing. Full model has 42.6B parameters (83.75 GB), each expert being 150MB in size. 
(2) Phi-tiny-MoE-Instruct \cite{microsoft2025phitinymoeinstruct}, a smaller variant from the PhiMoE model family with 3.8B params (7.51 GB in size). (3) Qwen3-30B-A3B \cite{yang2025qwen3} has 48 layers with a total of 128 experts (28.64 GB in size). 
Moving forward, we address them as PhiMoE, PhiMoE-tiny and Qwen3-MoE respectively.


\noindent\textbf{Workload}
We target long context understanding tasks such as QA, retrieval based tasks, MCQ etc. where prefill typically has a larger context and decode generates a few tokens; thus our primary focus is targeted towards ensuring \textit{prefill efficiency}.
We evaluate three representative datasets using standard few-shot prompting \cite{team2025kimi}: (a) HellaSwag \cite{zellers2019hellaswag}, a commonsense reasoning benchmark with narrative contexts and multiple-choice completions; (b) BoolQ \cite{clark2019boolq}, a QA dataset with yes/no answers; and (c) RULER \cite{hsieh2024ruler}, a long-context benchmark focused on retrieval and aggregation over extended inputs with short outputs.

\begin{table}
    \centering
    
    \caption{Comparing targets.}
    \vspace{-2mm}
\includegraphics[width=0.7\linewidth]{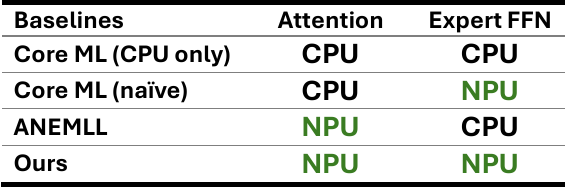}
    \Description{Comparison table of baseline systems evaluated against NPUMoE, summarizing their execution targets and high-level design differences.}
    \label{fig:b}
\end{table}
\noindent\textbf{Comparing targets}
We compare \sys{} with 3 baselines, all based on Core ML  to maintain consistency. They differ in where attention and expert FFNs execute. \autoref{fig:b} summarizes the placement policy for each system:
(1) \textit{Core ML (CPU)}: executes all operators on CPU.
(2) \textit{CoreML (na\"{\i}ve)}: default CoreML scheduling is used; expert FFN can be dispatched to NPU while control and unsupported paths remain on CPU. At runtime, Core ML decides if experts can run on NPU.
(3) \textit{ANEMLL} \cite{anemll}: library closest to ours that accelerates dense transformer LLMs for NPU. We preserve its original backbone to run dense transformer blocks on NPU and let the rest execute on CPU.

\sys{} represents our system with all our optimizations enabled. 
We enable chunked prefill following \cite{agrawal2023sarathi}. ANE is optimized for FP16 compute \cite{ane_number} and dequantizes INT8 to FP16 before computation \cite{singh2026m4ane}.
Unless otherwise noted, all comparisons use the same model checkpoint, tokenizer, FP16 precision, and real routing traces from workload. 

\noindent\textbf{Offline Calibration} 
We profile and evaluate on separate randomized subsets of our test datasets. Accuracy is evaluated by (1) prediction accuracy on unseen inputs and (2) cross-dataset validation. Both are measured using per-layer median Overlap@K \cite{li2025bild} between offline-profiled and runtime routed experts. 
Accuracy reported in \S\ref{sec:as}.

\noindent\textbf{Hyperparameters}
We vary (1) expert token capacity $\in\{32, 64, 128\}$, (2) group size $G\in\{4,8\}$, (3) prompt length $P\in\{1024,4096\}$, and (4) prefill chunk size $\in\{256,512,1024\}$. 

\noindent\textbf{Performance metrics}
We primarily optimize and focus on evaluating prefill performance.
We report three metrics:
(1) Runtime latency, measured in TTFT (time-to-first-token);
(2) Energy efficiency, measured as energy per token (EPT); calculated for N tokens as:
\[
\mathrm{EPT}=\frac{E_{\mathrm{compute}}+E_{\mathrm{comm.}}}{N_{\mathrm{tokens}}},
\]
$E_{\mathrm{compute}}$ means energy spent in compute by CPU, GPU (if active) and ANE combined. $E_{\mathrm{comm.}}$ measures energy spent in data movement and storage during inference.
(3) Reduction in CPU usage, measured by CPU cycles consumption. We use CPU cycles as an offload proxy to account for NPU utilization; fewer CPU cycles indicate that more compute is offloaded from CPU to NPU.

We also report end-to-end results covering both prefill and decode. According to our benchmark workloads, the average decode length is 8 tokens.
All device energies are collected with Zeus \cite{zeus_apple_silicon}.
Accuracy is measured with the LM Evaluation Harness \cite{eval-harness}.
Lower EPT, lower TTFT, and lower CPU cycle usage indicate better efficiency.

\begin{figure}
    \centering
    \includegraphics[width=0.9\linewidth]{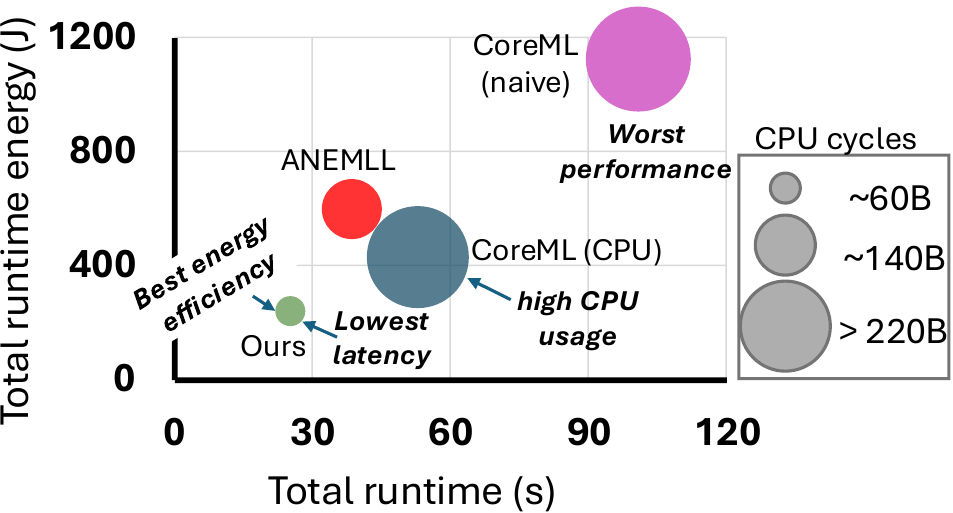}
    \caption{Trade-off analysis of latency, energy consumption, and CPU cycle usage for PhiMoE.
    \sys{} achieves the optimal pareto frontier compared to all baselines.}
    \Description{Scatter plot comparing latency, energy, and CPU-cycle usage for PhiMoE configurations, showing NPUMoE on the Pareto frontier against all baselines.}
    \label{fig:pareto}
\end{figure}

\subsection{Overall Performance}
Across all configurations, \sys{} consistently provides the best overall performance-efficiency trade-off, as shown in \autoref{fig:pareto}.
For PhiMoE on M2 Ultra, \sys{} achieves 1.32x–5.55x lower latency, 1.81x–7.37x higher energy efficiency, and 1.78x–5.54x lower CPU-cycle usage than the baselines across real routing traces with varying prompt lengths and chunk sizes during prefill (\S\ref{sec:rl}--\S\ref{sec:ane_util}).
This trend also holds for PhiMoE-tiny and Qwen3-MoE. Their speedups range from 1.11x to 1.87x and from 1.19x to 3.86x, while energy-efficiency gains range from 1.70x to 2.68x and from 1.84x to 3.89x, respectively (\S\ref{app:extended_eval}).
Accuracy is consistently maintained with <1.1\% degradation on our test datasets (ref. \autoref{tab:acc}). Thus we do not trade efficiency for performance. 


\subsection{Runtime Latency}
\label{sec:rl}

\begin{figure}
    \centering
    \includegraphics[width=\linewidth]{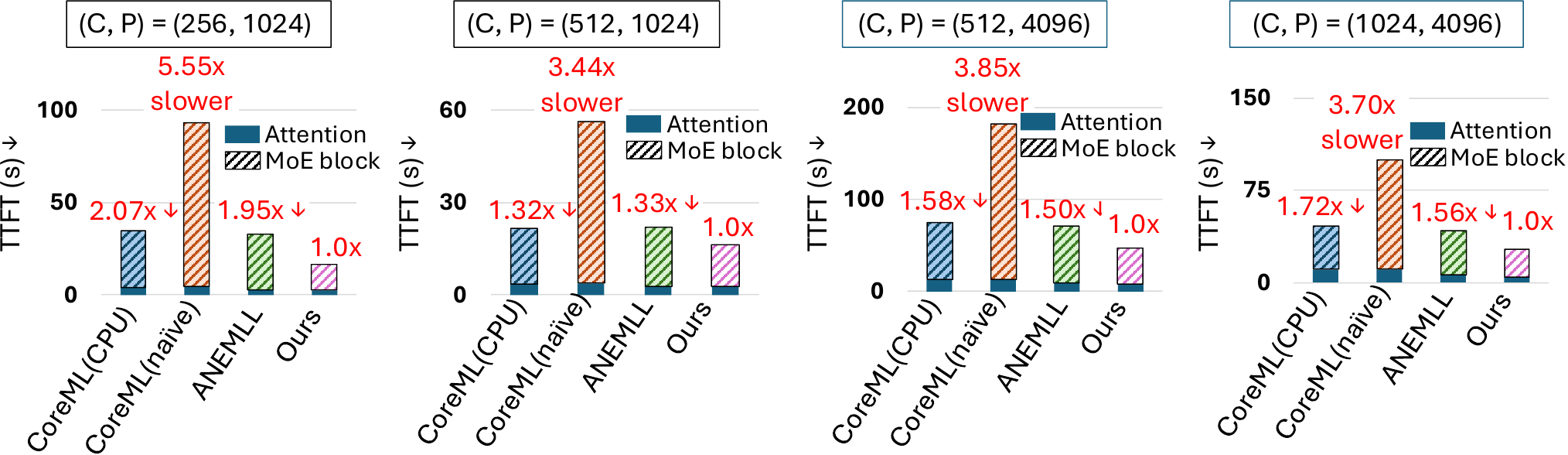}
    \caption{Prefill latency (TTFT) breakdown for PhiMoE on M2 Ultra across varying prompt lengths (P) and prefill chunk sizes (C), split into Attention and MoE components. Expert FFN takes up over 86\% of TTFT, indicating that expert execution dominates prefill latency.} 
    \Description{Stacked bar charts of PhiMoE prefill latency across prompt lengths and chunk sizes, separating attention and MoE costs and showing that expert execution dominates TTFT.}
    \label{fig:pl}
\end{figure}
\noindent\textbf{Prefill latency.} 
\sys{} consistently achieves the lowest prefill latency across all evaluated prompt lengths and chunk settings, outperforming all baselines by 1.32x–5.55x (\autoref{fig:pl}). At prompt length 1024, it reduces TTFT by 2.07x and 1.32x over CoreML(CPU), 5.55x and 3.44x over CoreML(naïve), and 1.95x and 1.33x over ANEMLL at chunk sizes 256 and 512, respectively. At prompt length 4096, \sys{} remains the fastest configuration, delivering 1.58x, 3.85x, and 1.50x lower latency at chunk 512, and 1.72x, 3.70x, and 1.56x lower latency at chunk 1024 over CoreML(CPU), CoreML(naïve), and ANEMLL, respectively. 

\noindent\textit{Breakdown:} \autoref{fig:pl} shows that TTFT reduction is driven mainly by the MoE block, while attention latency increases by $\sim$27.6\% depending on NPU or CPU backend. 
CPU-NPU dispatch can account for >60\% time as evident from the MoE block runtime of CoreML(naïve).

\noindent\textbf{End-to-end latency.}
\sys{} achieves the largest end-to-end gains on our 
long-context workloads with short outputs. 
Results show up to 3.86x speedup over CoreML(naïve), 1.26x over CoreML(CPU), and 1.19x over ANEMLL.


\begin{figure}[t]
    \centering
    \includegraphics[width=\linewidth]{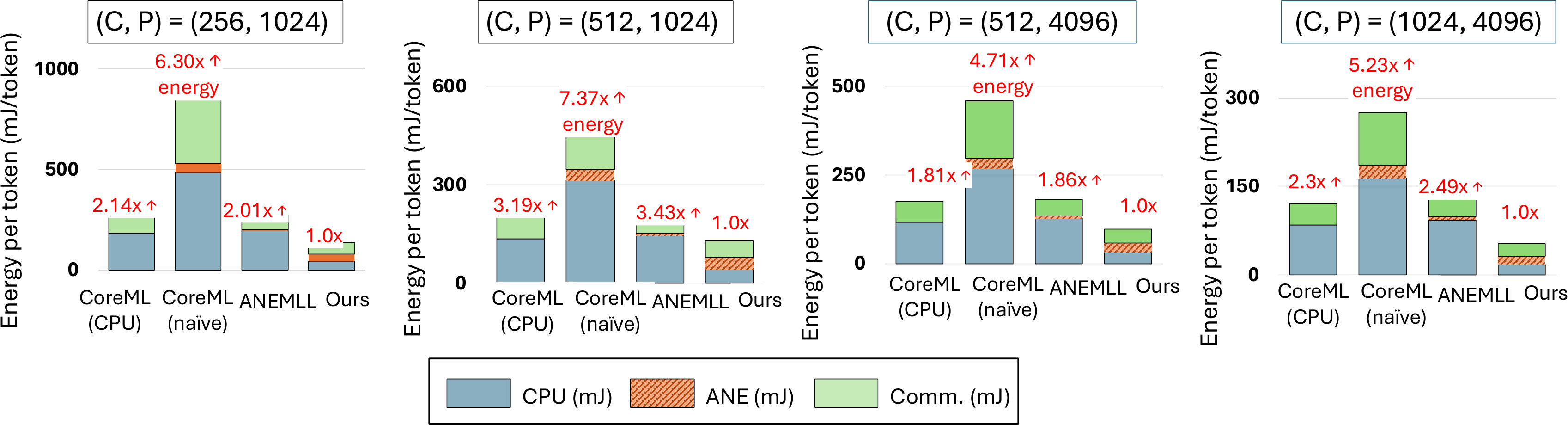}
    \caption{Energy consumption breakdown per token across varying prefill chunk sizes (C) and prompt lengths (P) for PhiMoE on the M2 Ultra. \sys{} consistently demonstrates the highest energy efficiency across all configurations.}
    \Description{Bar chart showing per-token energy across prompt lengths and chunk sizes for PhiMoE, with NPUMoE consistently using the least energy among the compared systems.}
    \label{fig:e}
    
\end{figure}
\subsection{Energy Efficiency}
\label{sec:energy_efficiency}
\noindent\textbf{Prefill.}
\autoref{fig:e} reports energy per token. 
Across all chunk sizes and prompt lengths, our system achieves the lowest energy per token. Relative to our design, CoreML (CPU), CoreML (naïve), and ANEMLL consume 1.81x–3.19x, 4.71x-7.37x, and 1.86x–3.43x higher energy per token, respectively. 

Our gains persist across both short and long prompts and remain robust as chunk size increases.
For short prompts (length=1024), the largest advantage appears at chunk size 512, where CoreML (naïve), CoreML (CPU), and ANEMLL consume 7.37x, 3.19x, and 3.43x more energy per token than ours, respectively.
For longer prompts, ours reduces energy per token by 4.71-5.23x over CoreML (naïve), 1.81-2.30x over CoreML (CPU), and 1.86-2.49x over ANEMLL.
Overall, our method maintains a low and stable energy cost of roughly 52–137 mJ/token. Notably, CoreML (naïve) spends 4.44x-5.69x more energy on data communication alone than ours, underscoring the overhead of naïve CPU–NPU coordination. 




\noindent\textbf{End-to-end.} \sys{} achieves the best efficiency with 3.89x lower energy than CoreML(naïve), 1.84x lower than CoreML(CPU), and 1.95x lower than ANEMLL end-to-end on our test datasets. 

\begin{figure}
    \centering
    \includegraphics[width=\linewidth]{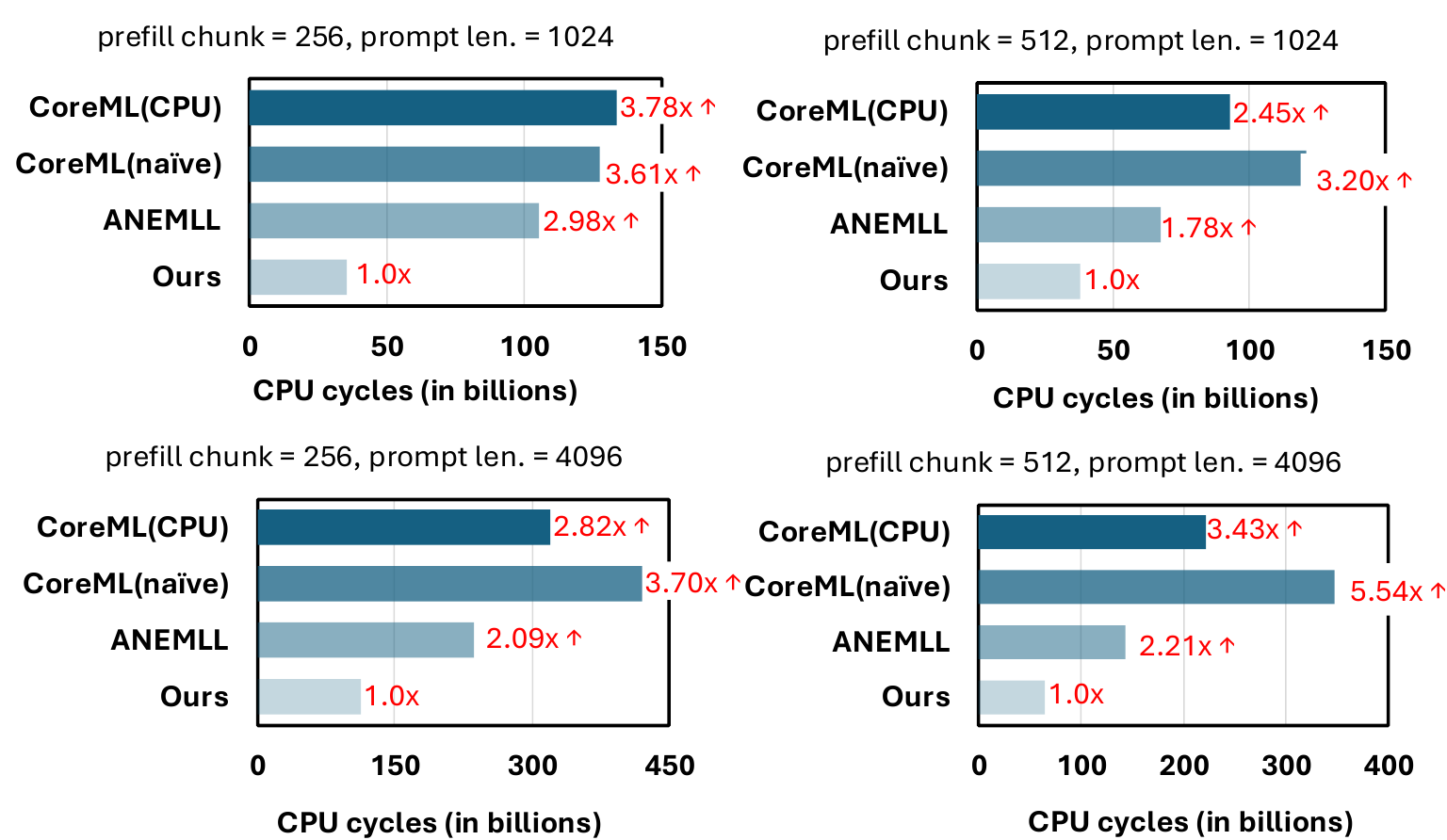}
    \caption{Total CPU cycles consumption across prefill workloads for PhiMoE on M2 Ultra (Lower is better). Reduced usage indicate effective offloading of compute to NPU.}
    \Description{Plot of total CPU-cycle usage across PhiMoE prefill workloads, showing that NPUMoE reduces CPU work by offloading more computation to the NPU.}
    \label{fig:cpu_cycles}
\end{figure}
\subsection{Reduction in CPU usage}
\label{sec:ane_util}
\noindent\textbf{Prefill.}
\autoref{fig:cpu_cycles} shows that our method consistently requires fewer CPU cycles than all other baselines.
For prompt length 1024, our method reduces CPU cycles by 3.78x, 3.61x, and 2.98x at prefill chunk 256, and by 2.45x, 3.20x, and 1.78x at prefill chunk 512, relative to CoreML (CPU), CoreML (naïve), and ANEMLL, respectively. For longer prompts (4096), the same trend holds: at prefill chunk 256, we reduce CPU cycles by 2.09-3.70x, and at prefill chunk 512, by 2.21x-5.54x, against baselines. 

\noindent\textbf{End-to-end.}
\sys{} consistently maintains the lowest total CPU cycles consumption, reducing count by 2.64x-3.75x over CoreML (CPU), 3.17x-4.67x over CoreML (naïve), and 2.01x-2.99x over ANEMLL across all test workloads.
Thus, our design effectively shifts compute off the CPU, as proven from reduced CPU cycles usage, increasing NPU utilization. 





\begin{table}
    \centering
    
    \caption{Inference accuracy of \sys{} on PhiMoE across four long-context workloads shows negligible degradation.}
    \includegraphics[width=\linewidth]{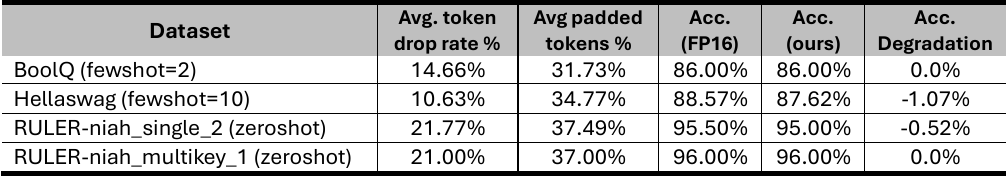}
    \Description{Table of evaluation accuracy across four long-context workloads showing that NPUMoE matches the reference model with negligible quality loss.}
    \label{tab:acc}
\end{table}

\subsection{Inference Accuracy}
\sys{} incurs negligible ($\sim$1\%) accuracy loss even under $\sim$10-20\% token drop rate (ref. \autoref{tab:acc}). 
Only low importance tokens are dropped, so accuracy is largely preserved: tokens still propagate via attention and residuals and are re-routed in later layers.
Prior work demonstrates that aggressively dropping tokens has negligible impact on accuracy \cite{zhang2025trimtokenator} and, by filtering out noisy or redundant data, can actually improve overall model performance \cite{yang2025topv}. 

\sys{}'s efficiency does not come from reducing effective work through token dropping since it still pads roughly 31.7\%-37.5\% of total tokens to satisfy static design constraints.
Both RULER tasks maintain 96.0\% accuracy despite the highest drop and padding rates, showing strong robustness for long-context retrieval workloads.


\subsection{Detailed evaluation of a Single MoE Block}
\label{sec:single_moe_block}

\begin{table}
    \centering
    \caption{MoE block runtime breakdown (in ms) for a single layer of PhiMoE on M2 Ultra (chunk=256, prompt len.=1024). Per-block latency is dominated by expert FFN and scatter/write-back cost, while router, padding etc. overheads remain negligible.}
\includegraphics[width=\linewidth]{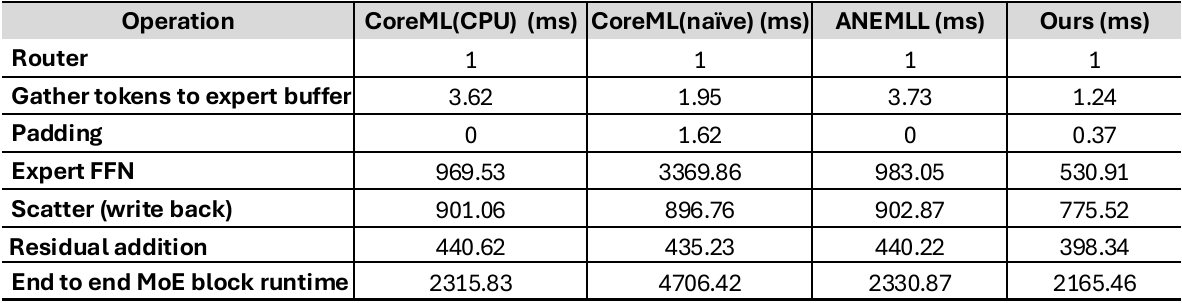}
    \Description{Table breaking down single-layer MoE block runtime into routing, packing, expert execution, and write-back components, with expert FFN and scatter dominating latency.}
    \label{tab:ob}
\end{table}

\noindent\textbf{Per operation runtime breakdown analysis.}
\autoref{tab:ob} illustrates runtime of each operation inside an MoE block. 
\sys{} reduces expert FFN time by $\sim$40–50\% and moderately improves scatter time by $\sim$10–15\%, yielding an overall speedup of 5–10\% over CoreML (CPU) and ANEMLL, and 54\% over CoreML (naïve). \textit{This indicates that system-level gains come primarily from improving expert parallelism and reducing data movement.}



\noindent\textbf{Varying per expert input token capacity.}
\begin{table}[t]
    \centering
    
    \caption{Per-token latency \& energy across expert grouping, capacity and execution mode for PhiMoE.}
    \includegraphics[width=\linewidth]{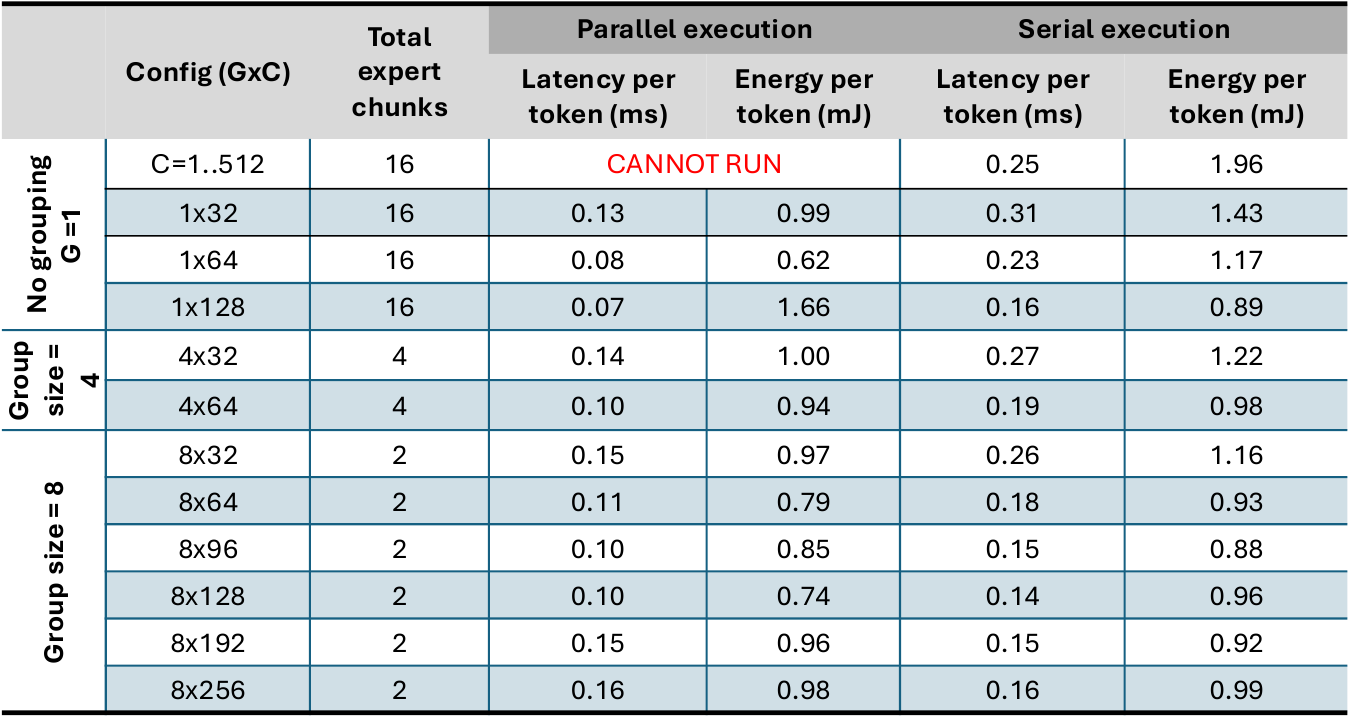}
    \Description{Table comparing per-token latency and energy for different expert grouping sizes, capacities, and execution modes, highlighting the trade-offs among them.}
    \label{fig:ep}
\end{table}
Increasing per-expert input token capacity improves latency and energy efficiency, although this benefit saturates quickly due to padding and graph launch overhead (ref. \autoref{fig:ep}).
Without grouping, latency drops by 1.94x from 0.31 ms/token at 1x32 to 0.16 ms/token at 1x128; energy efficiency improves by 1.22x-1.61x. 
\textit{With grouping, the benefit is larger}: latency improves by 1.44x at 8x64 and 1.73x at 8x96 over 8x32 at group size=8; energy efficiency improves by 1.25x-1.32x.



\noindent\textbf{Expert parallelism.} \autoref{fig:ep} shows the impact of expert parallelism on latency and energy. NPU supports limited concurrent compute graphs; exceeding this limit, e.g. many enumerated shapes, can trigger CPU fallback. Such as, running all 16 experts in parallel for token range 1 to 512 is infeasible, while serial execution is 1.4x-2.9x slower. \textit{Grouping mitigates this trade-off by reducing expert chunks and avoiding NPU queue serialization and CPU fallback.}


\noindent\textbf{Impact of group size on dispatch.}
Larger groups reduce dispatch overhead by lowering the number of expert graph executions: this effect is especially visible on NPUs where many small graphs amplify launch and synchronization cost (\autoref{fig:ep}). \textit{At the same per-expert capacity, throughput improves with group size}: for capacity 32, latency drops from 0.31 ms/token at {1x32} to 0.27 at {4x32} (1.15x) and 0.26 at {8x32} (1.19x); for capacity 64, it falls from 0.23 at {1x64} to 0.19 at {4x64} (1.21x) and 0.18 at {8x64} (1.28x). Energy per-token results in \autoref{fig:ep} indicates that larger groups create denser workloads and improve backend utilization.

\subsection{Ablation Study}
\label{sec:as}
\begin{figure}
    \centering
    \includegraphics[width=0.95\linewidth]{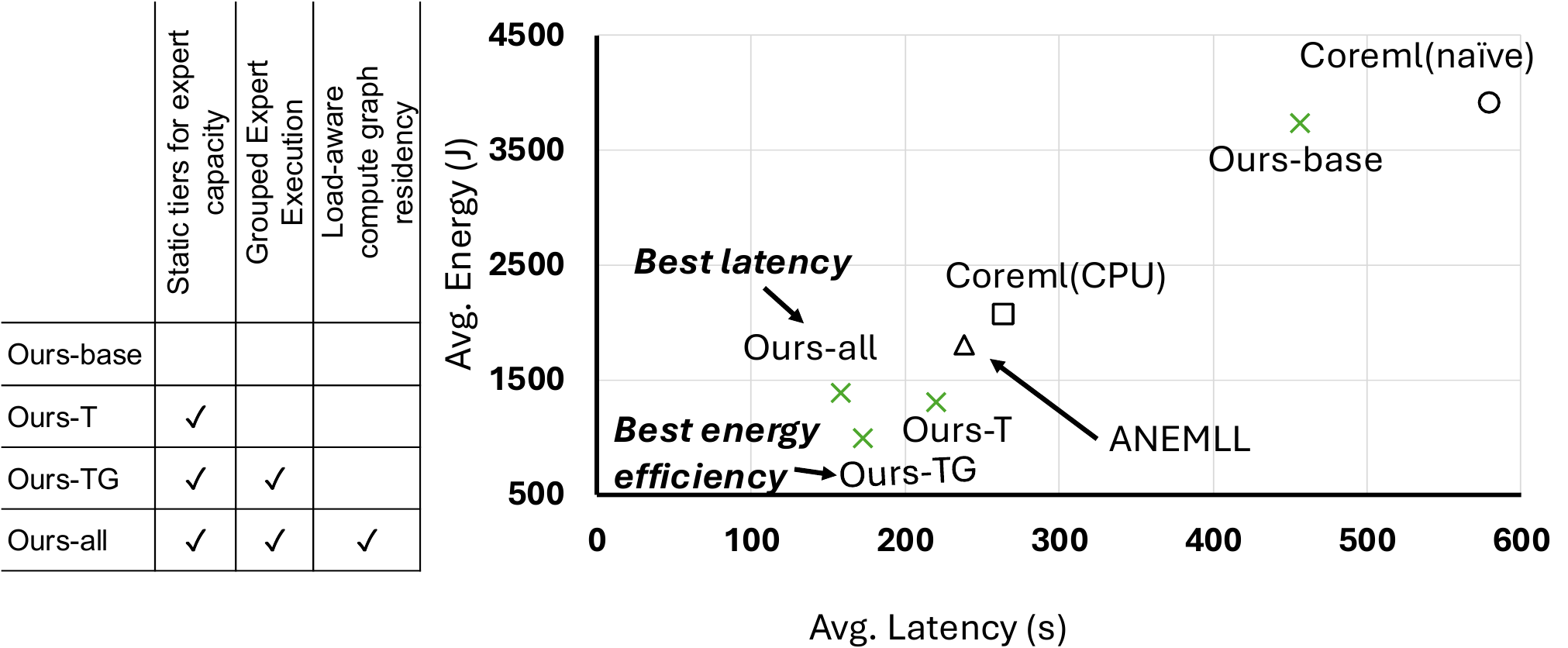}
    \caption{Impact of our key techniques.}
    \Description{Ablation summary showing the contribution of the main NPUMoE techniques to latency, energy efficiency, and CPU-cycle reduction.}
    \label{fig:kt}
\end{figure}
\noindent\textbf{Effectiveness of optimizations.} 
To isolate the contribution of each component, we also evaluate the following ablations:
(1) \noindent \textit{Ours-base}, which uses operator partitioning with a naïve one-graph-per-expert design and fixed expert shape;
(2) \noindent \textit{Ours-T}: Ours-base with static tiers for expert capacity;
(3) \noindent \textit{Ours-TG}: Ours-T with grouped expert execution; and
(4) \noindent \textit{Ours-all}: Ours-TG with load-aware expert compute residency.
\textbf{Our three techniques make a significant contribution to the overall improvement.}
\autoref{fig:kt} highlights how each technique contributes to the latency-energy tradeoff. Directly offloading prefill workloads to NPU without any of our techniques (Ours-base) gives the worst performance.
Ours-all achieves the best latency, 2.89x faster than Ours-base and 1.09x faster than Ours-TG, while Ours-TG gives the best energy efficiency at 998.4, using 3.74x less prefill energy than Ours-base and 1.31x less than Ours-T. Both grouped variants outperform the rest 
, indicating that per-dispatch overhead from 16 separate prediction calls is a major cost.

\noindent\textbf{Impact of padding.}
On our test workload, on average, 35.35\% of all computed tokens were zero-padded. Padding introduces 1.25-1.36x higher CPU cycles and
27–29\% higher runtime depending on device config. Thus, increasing expert capacity indiscriminately is not beneficial. 



\noindent\textbf{Does graph eviction happen?} Load and cache operation is done once at compile time; unless unified memory RAM is overloaded compute graph stays in place.

\noindent\textbf{Which prefill chunk size to choose?} 
Chunk selection is workload dependent. 
For shorter prompts (length=1024), increasing chunk size from 256 to 512 mainly improves energy, raising our advantage from 2.01x-6.30x to 3.19x-7.37x across baselines. For 4096-token prompts: increasing chunk size from 512 to 1024 widens the energy gap to 2.30x-5.23x, maintains the lowest TTFT at 1.44x-3.70x, and improves CPU-cycle reduction to 2.21x-5.54x. Larger chunks are more effective for long-context prefill, upto its upper bound \cite{agrawal2023sarathi}.



\begin{figure}
    \centering
    \includegraphics[width=0.95\linewidth]{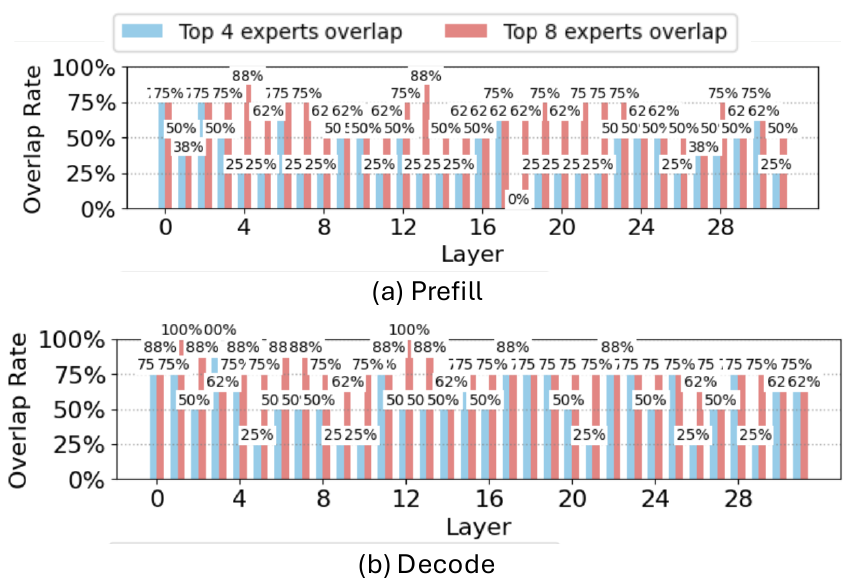}
    \caption{
    Cross-dataset validation for offline calibration. Overlap@K for top experts on BoolQ and HellaSwag datasets exhibit consistent overlap across dominant experts.}
    \Description{Cross-dataset comparison of top-ranked experts showing that the most active experts overlap substantially across BoolQ and HellaSwag, supporting offline calibration.}
    \label{fig:ec}
    \vspace{-3mm}
\end{figure}

\noindent\textbf{Calibration accuracy.} 
For prediction accuracy on unseen inputs, 
the mean overlap is 86\%/94\% for top-4/top-8 experts in prefill. 
Across datasets, PhiMoE achieves 65.2\% overlap in prefill and 77.0\% in decode, with many layers reaching near 100\% overlap (ref. \autoref{fig:ec}). 
Overall, offline calibration is effective at predicting dominant runtime experts; stability improves with larger K.

\noindent\textbf{Comparison to GPU.} GPU is 1.26x-5.4x faster, but uses 1.87x-4.48x more energy and draws 5.65x-9.41x higher power per run (Appendix \ref{app:extended_eval}). 
\noindent\textit{Note.} In this work, GPU serves a throughput reference only as our objective is not to match peak GPU throughput, but to reduce CPU pressure and improve sustained on-device efficiency by offloading suitable MoE operators to NPU.

\noindent\textbf{Limitation.}
Our optimization primarily targets long-context, prefill-dominated workloads. In our current prototype, decode follows the same execution pipeline as prefill; decode is memory-bound and single-token generation offers limited parallelism, it does not amortize CPU–NPU coordination overhead as effectively as prefill. 
\section{Related work} 
\noindent\textbf{On-Device LLM Acceleration} 
has largely focused on dense transformer inference on mobile GPUs \cite{cao2025moe} and NPUs e.g. llm.npu\cite{xu2025fast}, Hybe \cite{moon2025hybe}, HeteroLLM \cite{chen2025heterollm}, sd.npu \cite{chen2025accelerating}, OpenPangu \cite{dai2026accelerating} etc. using techniques such as chunked prefill \cite{agrawal2023sarathi}, graph bucketing \cite{yin2025dynamic}, tensor partitioning \cite{chen2025characterizing} etc. 
to accommodate the limitations of GPU or NPU while addressing key deployment constraints such as limited memory, static-shape, and hardware-aware scheduling. But they do not address the runtime challenges of executing dynamic and sparse models on NPUs.

           
\noindent\textbf{MoE Execution on Heterogeneous Devices}
Existing MoE inference frameworks primarily target memory-constrained deployments \cite{zhou2025floe, kong2024swapmoe, cao2025moe, kamahori2025fiddler} by leveraging heterogeneous CPU and GPU setups through techniques such as expert caching \cite{cao2025moe,xue2024moe}, prefetching \cite{yu2025prescope}, and tensor migration \cite{jung2023deepum}. 
MoE inference on mobile NPUs is largely unexplored, few prior work exist but they target custom, datacenter-class accelerators and their serving stacks e.g. Huawei’s CloudMatrix-Infer \cite{zuo2025serving} and xDeepServe \cite{xiao2025xdeepserve}. While NPU scheduling, virtualization have been explored \cite{xue2023v10, xue2024hardware}, they do not target on-device models.
\sys{} is the first to accelerate on-device MoE inference through effective NPU offloading. 

\section{Conclusion}
We present \sys{}, a runtime engine that enables efficient MoE inference on Apple Neural Engine by adapting dynamic expert routing to NPU’s device specific constraints. By combining key techniques,
\sys{} improves NPU utilization and consistently reduces latency, energy, and CPU-cycle usage on long-context workloads. 
These results show that, despite the mismatch between MoE dynamism and NPU constraints, careful systems co-design can make mobile NPUs an effective target for sparse LLM inference.

\section{Acknowledgements}
The authors acknowledge the use of generative AI to improve grammar, light editing, proofreading and code debugging. All content and ideas remain the original work of the authors.
\bibliographystyle{ACM-Reference-Format}
\bibliography{main}
\let\balance\relax
\newpage
\appendix
\section{Appendix}
\label{app:extended_eval}
\subsection{Phimoe-tiny}
\begin{figure}[H]
    \centering
    \includegraphics[width=\linewidth]{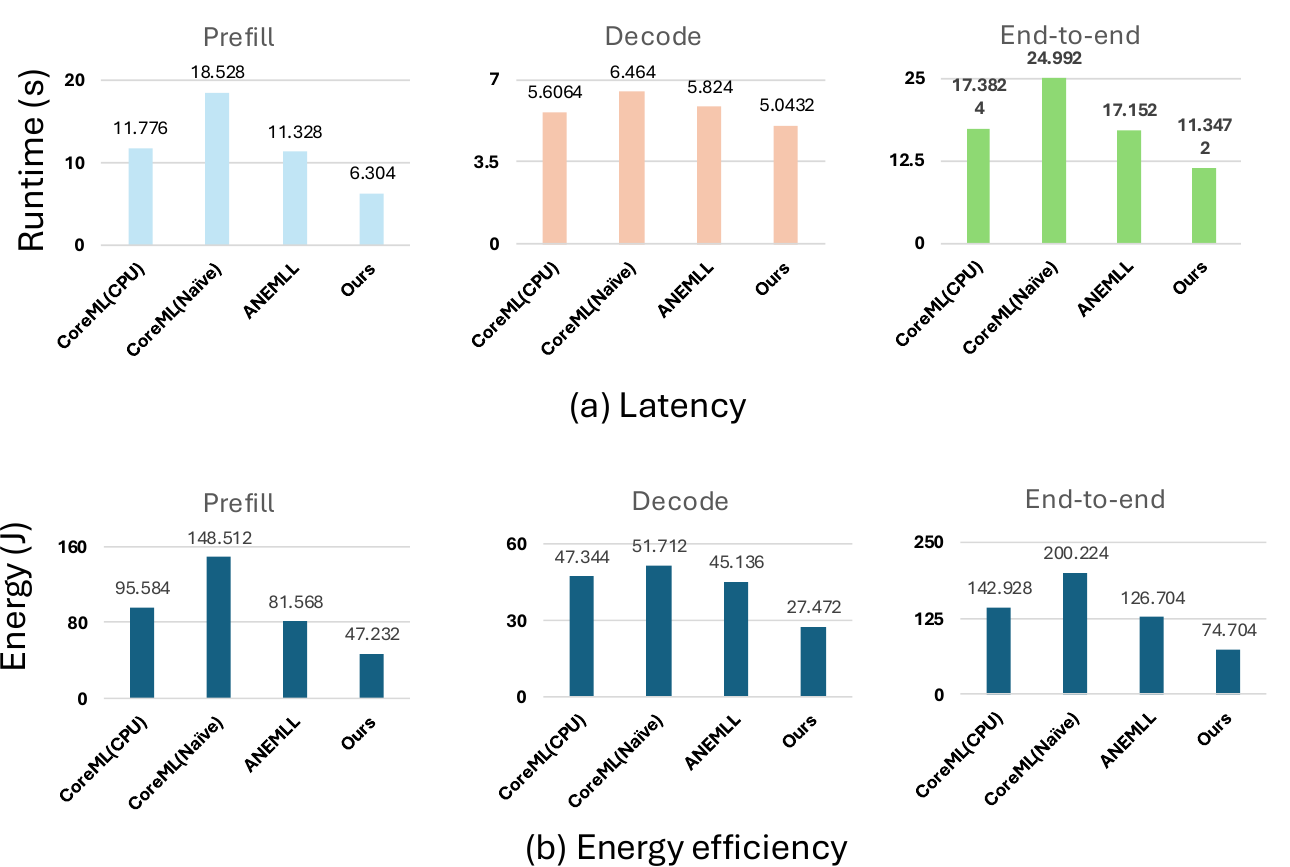}
    \caption{Runtime latency and energy efficiency of PhiMoe-tiny running on M2 MAX.}
    \Description{Plot showing runtime latency and energy efficiency of PhiMoe-tiny on M2 MAX.}
    \label{fig:placeholder}
\end{figure}

\subsection{Qwen3-MoE}
\begin{table}[h]
\centering
\small
\setlength{\tabcolsep}{4pt}
\caption{Latency breakdown of Qwen 30B on M2 Ultra. Speedup is relative to ours.}
\label{tab:lat}
\resizebox{0.9\linewidth}{!}{%
\begin{tabular}{@{}lcccc@{}}
\toprule
Mode & Wall (s) & Exp (s) & Attn (s) & Speedup \\
\midrule
ours      & 0.572 & 0.464 & 0.086 & 1.00x \\
coreml    & 1.139 & 0.978 & 0.095 & 1.99x \\
anemll    & 1.144 & 0.988 & 0.090 & 2.00x \\
\bottomrule
\end{tabular}
}
\end{table}

\begin{table}[H]
\centering
\small
\setlength{\tabcolsep}{3pt}
\caption{Energy efficiency of Qwen3-MoE compared to baselines, running on M2 Ultra.}
\label{tab:prefill_energy_chunk512}
\resizebox{\linewidth}{!}{%
\begin{tabular}{@{}lccccc@{}}
\toprule
Mode & Energy/run (J) & CPU (J) & ANE (J) & Comm. (J) & Energy Eff. \\
\midrule
Ours & 3.80  & 1.23 & 1.20 & 1.37 & 1.00x \\
coreml-only  & 12.67 & 9.66 & 0.00 & 3.01 & 3.33x \\
anemll       & 12.13 & 8.87 & 0.26 & 2.99 & 3.19x \\
\bottomrule
\end{tabular}
}
\end{table}

\subsection{Comparison with GPU backend}
\begin{table}[H]
\centering
\small
\setlength{\tabcolsep}{4pt}
\caption{Running a dense attention kernel with GPU vs NPU backend, averaged over 20 runs.}
\label{tab:attention_only_compare}
\resizebox{\linewidth}{!}{%
\begin{tabular}{@{}lrrl@{}}
\toprule
\textbf{Metric} & \textbf{GPU} & \textbf{NPU} & \textbf{GPU vs NPU} \\
\midrule
Runtime (ms) & 7.09 & 38.27 & GPU 5.40x faster \\
Energy (mJ) & 515.60 & 275.70 & GPU 1.87x $\uparrow$ energy \\
Avg power (W) & 66.80 & 7.10 & GPU 9.41x $\uparrow$ power draw \\
\bottomrule
\end{tabular}
}

\end{table}

\begin{table}[H]
\centering
\small
\setlength{\tabcolsep}{4pt}
\caption{Running attention+MoE block with GPU vs NPU backend.}
\label{tab:gpu_moe_compare}
\resizebox{\linewidth}{!}{%
\begin{tabular}{@{}lrrl@{}}
\toprule
Metric & GPU & Ours(NPU) & GPU vs.\ Ours(NPU) \\ \midrule
Runtime (s)    & 0.99 & 1.26 & GPU 1.26x faster \\ \hline
Energy/run (J) & 54.72 & 12.21 & GPU 4.48x $\uparrow$ energy \\ \hline
Avg power (W)  & 54.29 & 9.61 & GPU 5.65x $\uparrow$ power draw \\
\bottomrule
\end{tabular}
}

\end{table}

\subsection{Compute Graph}
\label{app:cg}
\begin{figure}[H]
    \centering
    \includegraphics[width=0.9\linewidth]{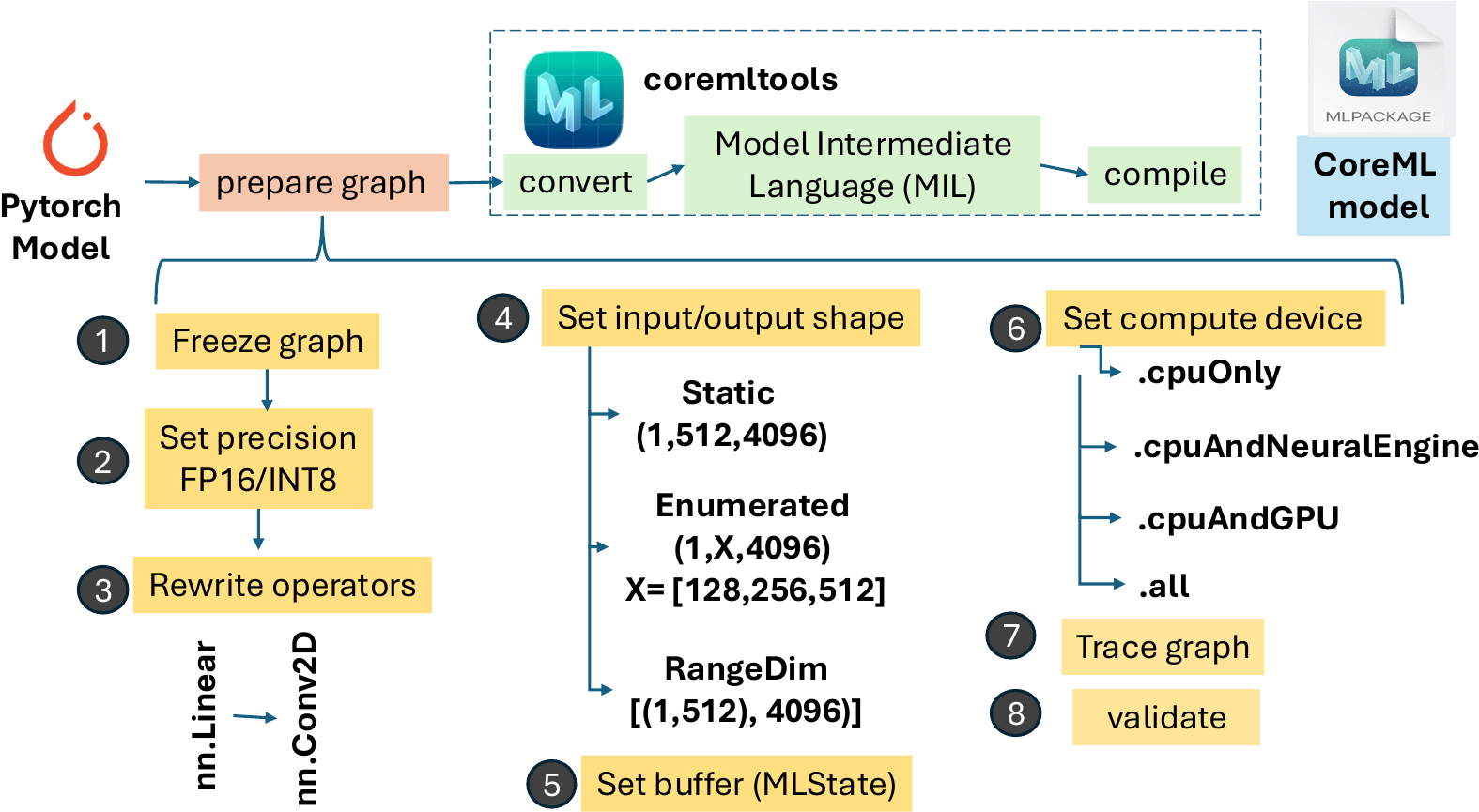}
    \caption{PyTorch to Core ML compute graph creation.}
    \Description{Pipeline diagram showing how PyTorch MoE components are converted into Core ML compute graphs for execution on Apple hardware.}
    \label{fig:conversion}
\end{figure}

\autoref{fig:conversion} shows the conversion pipeline from PyTorch models to NPU-compatible graphs.

\end{document}